\documentclass{article}

\usepackage{arxiv}

\usepackage[round]{natbib}

\usepackage{xcolor}
\usepackage[table]{xcolor}
\usepackage{amsthm}
\usepackage{amssymb}
\usepackage{amsmath}
\usepackage{tcolorbox}
\tcbuselibrary{listings}
\usepackage{thmtools} 
\usepackage{thm-restate}
\usepackage{tikz}
\usepackage{pgfplots}
\usetikzlibrary{plotmarks}
\usetikzlibrary{pgfplots.groupplots}
\usetikzlibrary{calc, trees, positioning, arrows, fit, shapes}
\usetikzlibrary{matrix}
\usepackage{multirow}
\usepackage{algorithm}
\usepackage{algorithmic}
\usepackage{hyperref}

\usepackage{stmaryrd}

\definecolor{forestgreen}{rgb}{0.13, 0.55, 0.13}

\newtheorem{definition}{Definition}

\begin{document}

\title{Retrieving Classes of Causal Orders with Inconsistent Knowledge Bases}
\author{\href{https://orcid.org/0009-0000-1323-4314}{\includegraphics[scale=0.06]{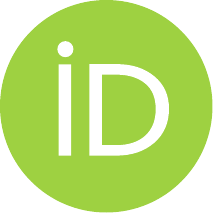}\hspace{1mm}Federico Baldo} \\ Sorbonne Université, INSERM,\\ Institut Pierre Louis d’Epidémiologie \\et de Santé Publique,\\ F75012, Paris, France 
\And
    \href{https://orcid.org/0009-0001-4277-8377}{\includegraphics[scale=0.06]{img/orcid.pdf}\hspace{1mm}Simon Ferreira} \\ Sorbonne Université, INSERM,\\ Institut Pierre Louis d’Epidémiologie \\et de Santé Publique,\\ F75012, Paris, France
\And
    \href{https://orcid.org/0000-0003-3571-3636}{\includegraphics[scale=0.06]{img/orcid.pdf}\hspace{1mm}Charles K. Assaad} \\ Sorbonne Université, INSERM,\\ Institut Pierre Louis d’Epidémiologie \\et de Santé Publique,\\ F75012, Paris, France
}
\maketitle





\begin{abstract}%
  Traditional causal discovery methods often depend on strong, untestable assumptions, making them unreliable in real-world applications.
In this context, Large Language Models (LLMs) have emerged as a promising alternative for extracting causal knowledge from text-based metadata, effectively consolidating domain expertise.
However, LLMs are prone to hallucinations, necessitating strategies that account for these limitations.
One effective approach is to use a consistency measure as a proxy of reliability. 
Moreover, LLMs do not clearly distinguish direct from indirect causal relationships, complicating the discovery of causal Directed Acyclic Graphs (DAGs), which are often sparse.
This ambiguity is evident in the way informal sentences are formulated in various domains. 
For this reason, focusing on causal orders provides a more practical and direct task for LLMs.
We propose a new method for deriving abstractions of causal orders that maximizes a consistency score obtained from an LLM.
Our approach begins by computing pairwise consistency scores between variables, from which we construct a semi-complete partially directed graph that consolidates these scores into an abstraction. 
Using this structure, we identify both a maximally oriented partially directed acyclic graph and an optimal set of acyclic tournaments that maximize consistency across all configurations.
We further demonstrate how both the abstraction and the class of causal orders can be used to estimate causal effects.
We evaluate our method on a wide set of causal DAGs extracted from scientific literature in epidemiology and public health.
Our results show that the proposed approach can effectively recover the correct causal order, providing a reliable and practical LLM-assisted causal framework.
\end{abstract}



\section{Introduction}
\label{sec:introduction}

Traditional causal discovery algorithms rely on observational (or interventional) data to uncover causal relationships.
To do so, they often make strong assumptions~\citep{Spirtes_2001,Glymour_2019,Peters_2017,Assaad_2022}, such as causal sufficiency and faithfulness.
The recent rise in popularity of Large Language Models (LLMs) offers a new tool to discover causal models~\citep{Long_2023,Long_2023a,Cohrs_2023,Vashishtha_2025,Kiciman_2024}.
Unlike traditional causal discovery methods, LLM-aided approaches operate on textual data, leveraging pre-collected knowledge encoded in their training data. 
However, LLMs have been frequently observed to provide ambiguous, inconsistent, or prompt-sensitive answers when queried about causal relationships~\citep{Zecevic_2023}.
Moreover, LLMs have limited capabilities when it comes to uncovering new knowledge or simply generating novel ideas, which undermines their applicability in the discovery of causal structures~\citep{Si_2025}

Most importantly, as noted in~\cite{Vashishtha_2025}, in natural language, direct and indirect causes are often conflated, making them difficult to distinguish.
This ambiguity is evident in informal statements across various domains, including philosophy, medicine, and epidemiology.
For instance, we commonly assert that a sedentary lifestyle causes type 2 diabetes, when in fact this link is fully mediated by obesity \citep{Li_2022}.
More generally, causal relations are frequently verbally expressed as a simple relationship:
"$X$ causes $Y$" or "$X$ affects $Y$" or "$X$ prevents $Y$", etc.
This oversimplification obscures the complex web of direct and indirect influences, including immediate "parents" and distant "ancestors" of a causal pathway.
In this paper, we argue that \emph{LLMs are not effective for discovering causal Directed Acyclic Graphs (DAGs), but rather allow us to retrieve abstractions of causal orders that can, under certain conditions, be used to estimate total effects}, or Average Treatment Effects (ATE).

In this context, we account for the inherent unreliability of LLMs, which often produce hallucinated or inconsistent responses.
We propose to quantify the reliability of the LLM and use such a metric as a heuristic to identify abstractions of causal orders.
Specifically, uncertainty in the LLM’s responses is captured through self-consistency, defined as the stability of its answers when queried multiple times about pairwise causal relationships.
By maximizing this consistency measure, we identify an abstraction in the form of a Semi-Complete Partially Directed Graph (PDG), which can be transformed into a dense Maximally oriented Partially Directed Acyclic Graph (MPDAG)~\citep{Perkovic_2020}.
We show that such an MPDAG can be used to establish a simpler criterion for the identifiability of the total effect in settings with a single treatment.
Additionally, based on the Semi-Complete PDG, we propose an exact method to derive all maximally consistent causal orders. 

Unlike traditional causal discovery methods~\citep{Spirtes_2001,Glymour_2019,Peters_2017,Assaad_2022}, our approach does not rely on faithfulness, or any parametric assumption, but only on acyclicity and causal sufficiency.
We view the LLM as an inconsistent knowledge base~\citep{Zheng_2024}, rather than an imperfect expert~\citep{Long_2023}, which can be used to rapidly access a large amount of information.
This perspective reflects the skepticism toward the ability of LLMs to discover new causal relationships~\citep{Zecevic_2023,Si_2025}.

\paragraph{Contributions}

\begin{itemize}
    \item We show how to derive a dense MPDAG from the knowledge provided by an LLM and how it can be used to establish a simpler criterion for identifiability of the total effect.
    \item We provide an effective algorithm to find a class of causal orders maximally consistent with an LLM. Such a method is based on a top-down search strategy that does not require any parametric assumptions or faithfulness.
\end{itemize}

The remainder of the paper is organized as follows: 
in Section~\ref{sec:related_work}, we review relevant literature related to LLM-aided causal discovery; 
in Section~\ref{sec:background}, we provide some of the basic formal notions used in the paper;
in Sections~\ref{sec:inconsistent_kb} and \ref{sec:mats}, we present the main contribution of the paper and details regarding the proposed method;
Sections~\ref{sec:experimental_results} and \ref{sec:discussion} conclude the paper with experimental results and final considerations.
Further details regarding the experimental setting, the implementation of the algorithm, and the proofs of the propositions and theorems presented in the paper are provided in the Appendix.

\section{Related Work}
\label{sec:related_work}

\paragraph{Causal Inference with Background Knowledge.}
The use of prior knowledge in causal discovery has been a long-standing research topic aimed at integrating domain-specific information to refine causal graphs.
\cite{Meek_1995} proposed a set of rules, known as Meek rules, to orient edges in a Completed Partially Directed Acyclic Graph (CPDAG) based on prior knowledge.
These rules allow us to refine the CPDAG by orienting edges while preserving the conditional independence encoded in the graph and the acyclicity constraint.
More recently, \cite{Maathuis_2015, Perkovic_2017,Perkovic_2017b,Perkovic_2020,Venkateswaran_2024} propose a generalization of identifiability by adjustment to abstraction of causal graphs, such as Partially Directed Acyclic Graphs (PDAGs), CPDAGs, and MPDAGs.

\paragraph{LLMs in Causal Discovery.}
\label{related_works:llm_aided_causal_discovery}
In causal discovery, LLMs have often been referred to as \textit{imperfect experts}~\citep{Long_2023,Vashishtha_2025}, since they are trained on vast amounts of textual data, including scientific literature. 
In this context, assuming we have a textual description of a set of variables \textemdash for instance, provided by a human expert \textemdash numerous works attempted to use LLMs to uncover causal DAGs~\citep{Kiciman_2024,Long_2023,Cohrs_2023,Vashishtha_2025,Jiralerspong_2024}.
However, LLMs tend to provide unreliable replies; most notably, they can hallucinate.
Concerns regarding their capability to effectively reason have been raised, as they might be just capturing verbal patterns without actually learning the underlying reasoning~\citep{Zecevic_2023}.
To tackle this problem, most of the approaches proposed until now quantify the reliability of these models.
As presented in~\cite{Cohrs_2025}, there are two primary approaches to compute uncertainties associated with the answers provided by LLMs: 
the first is to use the probabilities associated with the response tokens, i.e., the likelihood of the generated answer;
the second is to evaluate the \textit{consistency}, i.e., the self-coherence, of the LLM output when queried multiple times. 
In~\cite{Cohrs_2023}, authors propose an LLM-informed variant of the PC algorithm, where conditional independencies are detected by the language model.
In~\cite{Long_2023}, a pairwise prompt strategy is proposed to complete the orientation of edges in a CPDAG.
Each edge is associated with an uncertainty \textemdash based on the probabilities assigned to the response tokens \textemdash, then the Markov Equivalence Class (MEC) is refined through a Bayesian optimization process.
Lastly, in~\cite{Jiralerspong_2024}, the authors focus on an optimized query strategy to the LLMs aimed at reducing computational complexity.

\paragraph{LLMs and Causal Orders.}
\label{related_works:llm_and_causal_orders}
In~\cite{Vashishtha_2025}, the authors propose a method to estimate causal orders using LLMs.
As previously stated, the intuition is that LLMs can be more effective in identifying causal orders rather than causal DAGs, given the inherent ambiguity of causal relationships in natural language.
The method proposed estimates the topological order and identifies the total effect using the backdoor criterion~\citep{Pearl_2009}. 
To recover the causal order, the LLM is asked to provide a DAG for every possible triplet of variables, which are then aggregated to obtain a single causal order.

In this paper, we argue that identifying abstractions of causal orders rather than full causal DAGs presents a more direct task for LLMs.
However, unlike~\cite{Vashishtha_2025}, our approach focuses on:
1) using a pairwise prompt strategy to compute the self-consistency of the LLM,
2) identifying an abstraction of the maximally consistent causal orders, namely a fully connected Semi-Complete PDG,
and 3) refining the abstraction to obtain a dense MPDAG and a class of causal orders maximally consistent with the LLM. 
This design explicitly allows the inherent incongruity of the LLM to be reflected in the resulting causal representations, rather than forcing a single, potentially unreliable, structure.

\section{Background}
\label{sec:background}
\paragraph{Causal Graphs and Causal Orders}

A causal DAG, $\mathcal{G} = (\mathbb{V}, \mathbb{E})$, is defined by a set of nodes $\mathbb{V}$ and directed edges $\mathbb{E}$, where an edge $X_i \to X_j$ denotes a direct causal effect of $X_i$ on $X_j$  \citep{Pearl_2009,Spirtes_2001}.
In this context, the presence of confounding variables can lead to biased estimates of the causal effect.
To address this issue, assuming that all confounders are observed, we can use the backdoor criterion, which identifies sets of variables that, when controlled for, allow us to estimate the causal effect of $X_i$ on $X_j$ without biasing the estimate~\citep{Pearl_2009}.
The backdoor criterion is always satisfied if the conditioning set $\mathbb{Z}$ contains all the parents of $X_i$.
Given a causal DAG, we can define its causal order as follows:

\begin{definition} [Causal Order]
    Suppose a causal DAG $\mathcal{G}$. A causal order compatible with $\mathcal{G}$ is a bijective mapping $\pi: 
    \mathbb{V} \mapsto \{1, \ldots, d \}$ such that if $X_j$ 
    is a descendant of $X_i$ in $\mathcal{G}$, $X_i \succ_{\mathcal{G}} X_j$, then $\pi(X_i) < \pi(X_j)$, $\forall X_i, X_j\in \mathbb{V}$.
\end{definition}
Note that multiple causal orders can be compatible with the same causal DAG, as a single DAG may admit several valid topological orderings of its nodes.
A graph that admits a unique causal order does not permit ambiguity in the direction of the causal relationships between any pair of nodes.
We refer to these graphs as acyclic tournaments.

\begin{definition}[Acyclic tournament]
    An acyclic tournament is a DAG with one direct edge between every pair of distinct nodes.
\end{definition}

Acyclic tournaments provide a graphical representation that fully encodes a unique causal order.
Specifically, the direction of the edge between any two nodes directly reflects their relative position in the causal order.
Crucially, the backdoor criterion is always satisfied for causal orders by conditioning on all the parents of the treatment, assuming causal sufficiency~\citep{Pearl_2009,Vashishtha_2025}.
In the case of tournaments, the set of parents coincides with the set of predecessors in the causal order, eliminating the need to enumerate all paths between treatment and the outcome to identify a valid backdoor set.

\paragraph{Maximally Oriented Partially Directed Graphs (MPDAGs)}

In general, when causal sufficiency holds, and only observational data are available, if we do not make any parametric assumption, the best one can recover is a CPDAG~\citep{Spirtes_2001}.
A CPDAG is a partially directed graph that represents the equivalence class of all DAGs that are Markov equivalent to each other, meaning they encode the same conditional independences.
In the presence of background knowledge — for instance, provided by an expert — we can refine a CPDAG into an MPDAG~\citep{Perkovic_2017} by incorporating the additional edge orientations and applying the Meek rules to propagate their implications~\citep{Meek_1995}.
In the scope of this paper, we will mainly focus on the second Meek rule (R2), which enforces acyclicity in the graph, i.e., if orienting an undirected edge $X_i - X_j$ as $X_i \rightarrow X_j$ creates a directed cycle, then the edge must be oriented as $X_j \leftarrow X_i$.

The backdoor criterion cannot be directly applied to these abstractions, which include undirected edges; to this end, the generalized backdoor criterion~\citep{Maathuis_2015,Perkovic_2017} allows us to identify causal effects in CPDAGs. 
\begin{definition}[Generalized Backdoor Criterion]
\label{def:generlized_bd}
Let $\mathbb{X}_i$ and $\mathbb{X}_j$ be two disjoint sets of variables in a CPDAG $\mathcal{C}$. A set of variables $\mathbb{Z}$ satisfies the generalized backdoor criterion relative to $(\mathbb{X}_i, \mathbb{X}_j)$ if:

\begin{enumerate}
    \item $\mathbb{Z}$ contains no possible descendants of $\mathbb{X}_i$ in any DAG consistent with $\mathcal{C}$;
    \item Every path between $\mathbb{X}_i$ and $\mathbb{X}_j$ in $\mathcal{C}$ that contains an edge pointing into $\mathbb{X}_i$ is blocked by $\mathbb{Z}$.
\end{enumerate}
If $\mathbb{Z}$ satisfies this criterion, then the causal effect of $\mathbb{X}_i$ on $\mathbb{X}_j$ is identifiable and given by:
\begin{equation}
    \label{formula:generalized_adjustment}
    P(\mathbb{X}_j \mid do(\mathbb{X}_i = x)) = \sum_{z \in \mathbb{Z}} P(\mathbb{X}_j \mid \mathbb{X}_i = x, \mathbb{Z} = z) P(\mathbb{Z} = z).
\end{equation}
\end{definition}
The generalized backdoor criterion naturally extends to MPDAGs, which are enriched CPDAGs where all conditional independencies are preserved, and edge orientations are refined based on additional knowledge.
Whenever a backdoor set can be identified for an MPDAG, the total effect is identifiable and can be estimated using the standard adjustment formula~\eqref{formula:generalized_adjustment}.

\section{Inconsistent Knowledge Base}
\label{sec:inconsistent_kb}
The aim of this paper is to identify factual knowledge embedded in the LLMs from commonsense knowledge and established literature regarding cause-effect relationships in a specific domain.
We assume that these relationships do not directly indicate the presence of an edge in a causal DAG, but rather a causal ordering of the variables involved.
This knowledge is then used to identify abstractions of causal orders maximally compatible with the information provided by the LLM.
In this context, the LLM acts as a knowledge base~\citep{Zheng_2024} that: 1) has access to a large body of knowledge and 2) may provide incorrect responses, e.g., hallucinated responses. 
Measuring the reliability of an LLM's response is a well-established practice, which aims at quantifying the trustworthiness of the information provided by the model~\citep{Cohrs_2025}.
In this paper, we rely on self-consistency as a proxy of uncertainty \textemdash which has been shown to outperform other metrics, such as entropy, confidence elicitation, and token probabilities~\citep{Manakul_2023,Savage_2024}.

The goal is to define a consistency matrix encapsulating a measure of self-consistency of the LLM over pairwise causal relationships, which can be leveraged to identify maximally consistent causal orders.
Following the approach adopted in~\cite{Long_2023,Kadavath_2022}, we assume that we have a set of variables $X_1, ..., X_d$ associated with a set of textual descriptions, $\mu_1, ..., \mu_d$.
The consistency is the degree of agreement of the LLM when queried multiple times about the causal relationship between two variables, $X_i$ and $X_j$, with semantically equivalent queries.
The queries are generated by the LLM itself, which is asked to rephrase an initial sentence, such as \emph{"Is $\mu_i$ a cause of $\mu_j$?"}, into a set of semantically equivalent questions.
~To reduce the number of incorrect responses, we restrict the possible answers to a \texttt{Yes} or a \texttt{No} \textemdash more details in Appendix~\ref{appendix:graphs}.
The consistency score is then computed as the proportion of positive responses to the prompts.
Specifically, when an LLM is queried $n$ times, the consistency score for $X_i \succ X_j$  is calculated as:
$$C_{X_i \succ X_j} = \frac{1}{n}\sum_{k=1}^n r_k,$$
where $r_k$ is the response of the LLM to the $k$-th query, and $r_k = 1$ if the response is \texttt{Yes}, $r_k = 0$ if \texttt{No}.
~Note that $C_{X_i\succ X_j}$ and $C_{X_j\succ X_i}$ are computed independently of each other, i.e., there is no complementarity between the two.
As shown in Figure~\ref{fig:pipeline} (a), we can derive a consistency matrix, $C$, from the LLM's responses over a set of variables.
We formalize notions of consistency for a knowledge base as follows:

\begin{definition} [Consistent Knowledge Base]
    \label{def:consistent_kb}
    Given a causal DAG, $\mathcal{G}$, a knowledge base is \textit{consistent} if, for every pair of variables $X_i$ and $X_j$, such that $X_i \succ_{\mathcal{G}} X_j$ the consistency score $C_{X_i\succ X_j} \geq C_{X_j\succ X_i}$.
\end{definition}

\begin{definition} [Strictly Consistent Knowledge Base]
    \label{def:strictly_consistent_kb}
     Given a causal DAG, $\mathcal{G}$, a knowledge base is \textit{strictly consistent} if, for every pair of variables $X_i$ and $X_j$, such that $X_i \succ_{\mathcal{G}} X_j$ the consistency score $C_{X_i\succ X_j} > C_{X_j\succ X_i}$.
\end{definition}

\section{Retrieving Maximally Consistent Abstractions of Causal Orders}
\label{sec:mats}
The consistency matrix can be leveraged to construct a maximally consistent graph. 
Specifically, we can build a partially directed graph, $\mathcal{S}$, such that: a directed edge $X_i \rightarrow X_j$ is included in  $\mathcal{S}$ if $C_{X_i\succ X_j} > C_{X_j\succ X_i}$, and an undirected edge $X_i - X_j$ is included if $C_{X_i\succ X_j} = C_{X_j\succ X_i}$.
In Figure~\ref{fig:pipeline}, we illustrate how the consistency matrix can be used to derive such graph \textemdash in bold the consistencies that determine the orientation and the type of the edges.
Note that $\mathcal{S}$ can contain directed cycles, since there is no explicit mechanism preventing them.
This can be seen in Figure~\ref{fig:pipeline} (b), where the directed edges $W \rightarrow V$, $V \rightarrow Y$, and $Y \rightarrow W$ form a directed cycle.
We refer to a graph defined in this fashion as \emph{semi-complete partially directed graph}.

\begin{definition} [Semi-Complete Partially Directed Graph]
A semi-complete partially directed graph, or semi-complete PDG, is a dense graph in which every pair of distinct nodes is connected by exactly one edge, which may be either directed or undirected.
\end{definition}

\subsection{Maximally Consistent Semi-Complete Partially Directed Graphs}
\label{mats:semi-complete_directed_graph}

\begin{figure*}
    \centering
    \begin{tikzpicture}
        \matrix[matrix of nodes, nodes in empty cells, nodes={minimum size=0.5cm, font=\scriptsize, anchor=center}, column sep=0mm, row sep=0] (m) at (0, 0.3) {
        & Y & W & V & X & Z    \\
        Y   & -   & {\bf 0.6}  & 0.2 & 0.3 & {\bf 1.0} \\
        W   & 0.3 & -    & {\bf 0.7} & 0.4 & {\bf 0.4} \\
        V   & {\bf 0.7} & 0.5  & -   & 0.5 & {\bf 0.6} \\
        X   & 0.4 & {\bf 0.9}  & {\bf 0.9} & -   & 0.7 \\
        Z   & {\bf 1.0} & {\bf 0.4}  & {\bf 0.6} & {\bf 0.8} & -   \\
        };
        \node[text width=3cm, text centered, font=\footnotesize] (tauMap) at (0, -2) {(a)}; 

        \node[minimum height=0.5cm,minimum width=0.5cm] (Y) at (4.1, 2) {$Y$};
        \node[minimum height=0.5cm,minimum width=0.5cm] (X) at (5.1, -1) {$X$};
        \node[minimum height=0.5cm,minimum width=0.5cm] (V) at (3.1, 1) {$V$};
        \node[minimum height=0.5cm,minimum width=0.5cm] (Z) at (3.1, -1) {$Z$};
        \node[minimum height=0.5cm,minimum width=0.5cm] (W) at (5.1, 1) {$W$};

        \draw[->,>=latex] (W) -- (V);
        \draw[->,>=latex] (V) -- (Y);
        \draw[->,>=latex] (Y) -- (W);
        \draw[->,>=latex] (X) -- (V);
        \draw[->,>=latex] (Z) -- (X);
        \draw[->,>=latex] (X) -- (W);
        \draw[->,>=latex] (X) -- (Y);
        
        \draw[-] (Z) -- (Y);
        \draw[-] (Z) -- (V);
        \draw[-] (W) -- (Z);

        \node[text width=6cm, text centered, font=\small] (tauMap) at (4.1, -2) {(b)}; 

        \node[minimum height=0.5cm,minimum width=0.5cm] (Y1) at (12.1, 2) {$Y$};
        \node[minimum height=0.5cm,minimum width=0.5cm] (X1) at (13.1, -1) {$X$};
        \node[minimum height=0.5cm,minimum width=0.5cm] (V1) at (11.1, 1) {$V$};
        \node[minimum height=0.5cm,minimum width=0.5cm] (Z1) at (11.1, -1) {$Z$};
        \node[minimum height=0.5cm,minimum width=0.5cm] (W1) at (13.1, 1) {$W$};

        \draw[->,>=latex,ultra thick] (V1) -- (W1);
        \draw[->,>=latex] (V1) -- (Y1);
        \draw[->,>=latex] (Y1) -- (W1);
        \draw[->,>=latex] (X1) -- (V1);
        \draw[->,>=latex] (Z1) -- (X1);
        \draw[->,>=latex] (X1) -- (W1);
        \draw[->,>=latex] (X1) -- (Y1);
        \draw[->,>=latex,ultra thick] (Z1) -- (Y1);
        \draw[->,>=latex,ultra thick] (Z1) -- (V1);
        \draw[->,>=latex,ultra thick] (Z1) -- (W1);

        \node[text width=6cm, text centered, font=\small] (tauMap) at (8.1, -2) {(c)}; 


        \node[minimum height=0.5cm,minimum width=0.5cm] (Y2) at (8.1, 2) {$Y$};
        \node[minimum height=0.5cm,minimum width=0.5cm] (X2) at (9.1, -1) {$X$};
        \node[minimum height=0.5cm,minimum width=0.5cm] (V2) at (7.1, 1) {$V$};
        \node[minimum height=0.5cm,minimum width=0.5cm] (Z2) at (7.1, -1) {$Z$};
        \node[minimum height=0.5cm,minimum width=0.5cm] (W2) at (9.1, 1) {$W$};

        \draw[-,ultra thick] (V2) -- (W2);
        \draw[-,ultra thick] (Y2) -- (V2);
        \draw[-,ultra thick] (Y2) -- (W2);
        \draw[->,>=latex] (X2) -- (V2);
        \draw[->,>=latex] (Z2) -- (X2);
        \draw[->,>=latex] (X2) -- (W2);
        \draw[->,>=latex] (X2) -- (Y2);
        \draw[->,>=latex,ultra thick] (Z2) -- (Y2);
        \draw[->,>=latex,ultra thick] (Z2) -- (V2);
        \draw[->,>=latex,ultra thick] (Z2) -- (W2);

        \node[text width=6cm, text centered, font=\small] (tauMap) at (12.1, -2) {(d)}; 

    \end{tikzpicture}
    \caption{(a) Consistency Matrix. (b) Maximally consistent Semi-Complete PDG. (c) MPDAG derived from the Semi-Complete PDG. (d) Maximally consistent acyclic tournament. In bold edges modified with respect to the Semi-Complete PDG.}
    \label{fig:pipeline}

\end{figure*}

The maximally consistent semi-complete PDG, $\mathcal{S}$, holds specific properties based on the consistency of the knowledge base used to construct it.
Indeed, we can show that $\mathcal{S}$ is a dense MPDAG if the LLM is consistent (Definition~\ref{def:consistent_kb}). 

\begin{restatable}{proposition}{propositionone}
    \label{prop:1}
    The maximally consistent semi-complete PDG, $\mathcal{S}$, does not contain directed cycles if the consistency matrix is provided by a consistent knowledge base.
\end{restatable}

\begin{restatable}{proposition}{propositiontwo}
    \label{prop:2}
    The maximally consistent semi-complete PDG, $\mathcal{S}$, obtained from a consistent knowledge base and by applying R2 of the Meek rules, is an MPDAG.
\end{restatable}

Note that a dense MPDAG, i.e., with an edge between every pair of variables, has no v-structures, but only shielded colliders; this implies that we only require R2 for maximal orientation.
Unfortunately, we have no guarantees on the consistency of the LLM; thus, the semi-complete PDG may contain directed cycles.
In this case, we can still derive an MPDAG from $\mathcal{S}$ by unorienting directed cycles and enforcing acyclicity.

\begin{restatable}{proposition}{propositionthree}
    \label{prop:3}
    Assuming a semi-complete PDG, $\mathcal{S}$, containing directed cycles, $C_1,\ldots,C_m$, we can derive an MPDAG from $\mathcal{S}$ by:
    \begin{enumerate}
        \item Unorienting every edge in a cycle $C_k$, i.e., for every pair of nodes $X_i$ and $X_j$, such that $X_i \rightarrow X_j$ is part of a cycle $C_k$, we substitute the directed edge with an undirected one, $X_i - X_j$.
        \item Maximally orient undirected edges applying R2 of the Meek rules.
    \end{enumerate}
\end{restatable}

An application of Proposition~\ref{prop:3} is illustrated in Figure~\ref{fig:pipeline} (c), where the directed cycle involving nodes $Y$, $W$, and $V$ is unoriented, and then R2 is applied to orient additional edges.
Note also that the one-to-one correspondence between tournaments and orders does not hold for MPDAGs, which rather map to a set of partial orders.

\paragraph{Reasoning with Dense Maximally Oriented Partially Directed Graphs.}

As per~\cite{Maathuis_2015, Perkovic_2017b}, we can use MPDAGs to identify the total effect through the generalized backdoor criterion.
We show that a dense MPDAG, obtained from the maximally consistent Semi-Complete PDG, as per Proposition~\ref{prop:1}, \ref{prop:2}, and \ref{prop:3}, allows us to establish a simpler identifiability criterion for univariate treatments:

\begin{restatable}{proposition}{propositionfour}
    \label{prop:4}
    Given a dense MPDAG and the treatment variable $X_i$, the total effect is identifiable, using the adjustment formula~\eqref{formula:generalized_adjustment}, if and only if there are no undirected edges related to $X_i$.
\end{restatable}

This implies that it suffices to examine whether any undirected edges are incident on the treatment variable $X_i$ to determine if the total effect can be identified via adjustment.
Note that some nodes in $\mathcal{S}$ may be \emph{order invariant}, meaning that they are connected to every other node with undirected edges.
These nodes do not provide any information about the causal order and prevent identifiability.
\begin{restatable}{corollary}{corollaryone}
    \label{corollary:1}
    Given a dense MPDAG, if it contains an order invariant node, $X_i$, any causal effect between $X_i$ and every other variable is not identifiable.
\end{restatable}

The key assumption underlying the construction of a dense MPDAG from the consistency matrix is that the LLM provides reliable information about pairwise causal relationships.
While this assumption may not always hold, Proposition~\ref{prop:3} allows us to handle inconsistencies in the knowledge base by unorienting directed cycles. 
Alternatively, one could interpret cycles within $\mathcal{S}$ as potential errors, which can be addressed by applying corrective procedures that enforce acyclicity.

\subsection{Maximally Consistent Causal Orders}
\label{sec:maximal_acyclic_tournaments}
Orienting all edges while enforcing acyclicity and maximizing consistency over the semi-complete PDG, $\mathcal{S}$, allows us to identify a class of maximally consistent acyclic tournaments (MCAT).
Each acyclic tournament corresponds to a unique causal order, and vice versa (Section~\ref{sec:background}).
To establish a connection between semi-complete PDGs and acyclic tournaments, we introduce the notion of compatibility: 
\begin{definition}[Compatible Acyclic Tournament]
    \label{compatible_graph}
    Given a semi-complete PDG, $\mathcal{S}$, an acyclic tournament, $\mathcal{T}$, is said to be compatible with $\mathcal{S}$ 
    if:
    \begin{enumerate}
        \item $\mathcal{T}$ is obtained by reversing certain directed edges of $\mathcal{S}$ to produce an acyclic directed graph, while leaving all other edges unchanged.
        \item $\mathcal{T}$ contains an acyclic orientation of all undirected edges.
    \end{enumerate} 
\end{definition}

We can then identify a causal order that maximally aligns with the knowledge provided by the LLM by finding all MCATs compatible with the semi-complete PDG.

\begin{definition}[Maximally consistent acyclic tournament]
    Given a semi-complete PDG, $\mathcal{S}$, an acyclic tournament, $\mathcal{T}$, compatible with $\mathcal{S}$, is said to be maximally consistent if it maximizes the consistency score relative to all other acyclic tournaments
    compatible with $\mathcal{S}$. 
\end{definition}

In this context, if $\mathcal{S}$ contains directed cycles or undirected edges, we need to obtain an acyclic transformation. 
Finding such a transformation is a well-studied problem in graph theory, known as the Feedback Arc Set (FAS) problem~\citep{Karp_1972}. 
In its general form, the FAS problem consists of identifying a minimal set of edges whose removal eliminates all cycles in a directed graph.
The same set can be used to obtain an acyclic graph by reversing the edges instead of removing them~\citep{Barthelemy_1995}.
In our setting, we aim to find the acyclic transformation of $\mathcal{S}$ that maximizes the consistency score. 
To this end, we consider a weighted version of the FAS, seeking the minimal set of edges whose reversal achieves the highest score.
Instead of considering consistency scores as edges' weights, we define a cost score, $B_{i\rightarrow j}$, for each edge, $X_i \rightarrow X_j$, in $\mathcal{S}$:
\[
B_{i\rightarrow j} = \xi - C_{X_i \succ X_j} + C_{X_j \succ X_i}.
\]
where $\xi$ is the total consistency score of $\mathcal{S}$, calculated as the sum of the consistency scores of all edges.
The cost score $B_{i\rightarrow j}$ represents the cost of reversing the edge $X_i \rightarrow X_j$ in $\mathcal{S}$.
The goal is to find the solution to the FAS that maximizes the total cost score, which is equivalent to maximizing the total consistency score of the resulting acyclic tournament.
Thus, we can identify the set of edges, $A$, whose reversal leads to an MCAT by solving the following optimization problem:
\begin{equation}
    \label{eq:fas}
    A = arg \max_{\hat{A} \in \mathcal{F}_{\mathcal{S}}} \sum_{X_i \succ X_j \in \hat{A}} B_{i\rightarrow j} 
\end{equation}
where $\mathcal{F}_{\mathcal{S}}$ is the set of all FAS of $\mathcal{S}$.
The second step consists of orienting the undirected edges in $\mathcal{S}$.
Since undirected edges have equal consistency scores in both directions, we can consider all acyclic orientations of the undirected edges as equally consistent \textemdash for more details, see Appendix~\ref{appendix:implementation}.
In Figure~\ref{fig:pipeline} (d), we illustrate an MCAT compatible with the semi-complete PDG in Figure~\ref{fig:pipeline} (b).

\paragraph{Finding All Maximally Consistent Acyclic Tournaments.}
\label{sec:maximal_acyclic_tournaments_search}

The FAS maximizing the total cost score and compatible with $\mathcal{S}$ is not necessarily unique.
We can find all such sets by iteratively solving the weighted FAS while excluding previously found solutions.
Let $A$ be an optimal solution to \eqref{eq:fas}, meaning that there is an MCAT, $\mathcal{T} = (V, E_{\mathcal{T}})$, such that $E_{\mathcal{T}} = (E_{\mathcal{S}} \cap \mathbb{A}) \cup \mathbb{A}^{{T}}$.
We want to find all the acyclic tournaments $\mathcal{T}' = (V, E_{\mathcal{T}}')$ such that $E_{\mathcal{T}}' = (E_{\mathcal{S}} \cap \mathbb{A}') \cup \mathbb{A}^{'T}$ and $\mathbb{A}' \neq \mathbb{A}$, such that $A'$ is a solution to \eqref{eq:fas}.
This can be obtained by excluding subsets of edges of the optimal solution, $\mathbb{F} \in \mathcal{P}(\mathbb{A})$, from the search space, i.e., we will search through all solutions to the FAS where the edges in $\mathbb{F}$ are prohibited from being reversed.
If the exclusion of $\mathbb{F}$ leads to a suboptimal solution \textemdash non-maximal \textemdash then any other solution containing $\mathbb{F}$ will be suboptimal.
This allows us to prune the search space efficiently.
We denote this method as {\bf M}aximal Weighted {\bf A}cyclic {\bf T}ournaments {\bf S}earch, or MATS \textemdash more details about the procedure can be found in Appendix~\ref{appendix:implementation}.

\begin{restatable}{theorem}{theoremone}
    \label{thm:1}
    The MATS algorithm is sound, complete, and terminates.
\end{restatable}

\begin{restatable}{proposition}{propositionfive}
\label{prop:5}
    If the consistency matrix is provided by a consistent knowledge base, 
    the MATS algorithm is guaranteed to return a class of acyclic tournaments
    containing the true causal order.
\end{restatable}

\begin{restatable}{proposition}{propositionsix}
    \label{prop:6}
    If the consistency matrix is provided by a strictly consistent knowledge base, 
    the MATS algorithm is guaranteed to return the true causal order.
\end{restatable}

\paragraph{Reasoning with Classes of Causal Orders.}

If a treatment variable $X_i$ is part of a cycle in the Semi-Complete PDG, the total effect is not identifiable by deriving the associated MPDAG (Proposition~\ref{prop:3} and \ref{prop:4}).
In this case, we can estimate the total effect using the class of maximally consistent causal orders.
Given an estimated causal order, $\hat{\pi}$, we can use the predecessors of the treatment, $X_i$, to estimate the total effect over another variable, $X_j$, assuming that $\hat{\pi}(X_i) > \hat{\pi}(X_j)$ and causal sufficiency~\citep{Pearl_2009,Vashishtha_2025}.
This allows for efficient identification of the adjustment set without enumerating all backdoor paths from $X_i$ to $X_j$.
Moreover, we do not require the correct causal order to identify the total effect as long as all the predecessors are correctly ordered with respect to the treatment variable.
We can generalize this to every maximally consistent causal order in the recovered class.
In this context, two or more orders might share the same predecessors, e.g, if ${\hat{\pi}}(\mathbb{Z}) > {\hat{\pi}}(X_i) > {\hat{\pi}}(\mathbb{M}) > {\hat{\pi}}(X_j)$ and ${\hat{\pi}}(\mathbb{Z}) > {\hat{\pi}}(X_i) > {\hat{\pi}}(\mathbb{M}') > {\hat{\pi}}(X_j)$, meaning that the total effect is the same for both orders.
This implies that we can efficiently estimate the total effect once for multiple orders sharing the same backdoor set.
Ultimately, the total effect for an MPDAG is uniquely identifiable, whereas for a class of causal orders, we might have multiple estimations.


\section{Experimental Results}
\label{sec:experimental_results}

The code\footnote{\url{https://github.com/Federic0Bald0/MATS}} is designed to be compatible with OpenAI API and open-source LLMs available through the Ollama platform.
Results presented in Table~\ref{tab:results} are obtained using \texttt{gpt-4.1-nano}\footnote{https://platform.openai.com/docs/models/gpt-4.1-nano}, a fast and lightweight version of \texttt{gpt}. 
Additional results relative to \texttt{mistral:7b}\footnote{https://ollama.com/library/mistral} and \texttt{llama3.1:8b}\footnote{https://ollama.com/library/llama3.1} are available in Table~\ref{tab:results_llms_orders} of Appendix~\ref{appendix:additional}.
To minimize hallucinations and ensure a higher degree of consistency, all the experiments relative to LLMs have been conducted with \emph{temperature} set to $0.1$ \textemdash more details in Appendix~\ref{appendix:additional}.
Graphs are implemented using \texttt{igraph}, a \texttt{C++} library that offers an implementation of \texttt{ExactFAS} based on an integer programming formulation, which is guaranteed to yield an optimal result~\citep{Baharev_2021} \textemdash more details in Appendix~\ref{appendix:implementation}. 
Note that finding a FAS is NP-complete, even though for larger graphs, one could also resort to heuristic methods~\citep{Eades_1993}, also available on \texttt{igraph}.

\begin{table}
\centering
\scriptsize
\setlength{\tabcolsep}{2pt}
\begin{tabular}{cccccccccc}
\hline
\multicolumn{1}{|c|}{\multirow{2}{*}{\textbf{Graph}}} & \multicolumn{1}{c|}{\multirow{2}{*}{\textbf{PC}}}                                   & \multicolumn{1}{c|}{\multirow{2}{*}{\textbf{GES}}}                                  & \multicolumn{1}{c|}{\multirow{2}{*}{\textbf{NOTEARS}}} & \multicolumn{1}{c|}{\multirow{2}{*}{\textbf{BOSS}}}                                 & \multicolumn{1}{c|}{\multirow{2}{*}{\textbf{GRaSP}}}                                & \multicolumn{1}{c|}{\multirow{2}{*}{\textbf{GOBNILP}}} & \multicolumn{1}{c|}{\multirow{2}{*}{\textbf{Triplet}}}                              & \multicolumn{2}{c|}{\textbf{MATS (Our)}}                                                                                                                                  \\ \cline{9-10} 
\multicolumn{1}{|c|}{}                                & \multicolumn{1}{c|}{}                                                               & \multicolumn{1}{c|}{}                                                               & \multicolumn{1}{c|}{}                                  & \multicolumn{1}{c|}{}                                                               & \multicolumn{1}{c|}{}                                                               & \multicolumn{1}{c|}{}                                  & \multicolumn{1}{c|}{}                                                               & \multicolumn{1}{c|}{\textbf{MPDAG}}                                                 & \multicolumn{1}{c|}{\textbf{$\pi$}}                                                 \\ \hline
\multicolumn{1}{|c|}{\textbf{Asia}}                   & \multicolumn{1}{c|}{$3.66^{\pm 2.47}$}                                              & \multicolumn{1}{c|}{$1.75^{\pm 1.20}$}                                              & \multicolumn{1}{c|}{$4.08^{\pm 0.92}$}                   & \multicolumn{1}{c|}{$1.50^{\pm 0.87}$}                                              & \multicolumn{1}{c|}{$1.50^{\pm 0.87}$}                                              & \multicolumn{1}{c|}{$4.31^{\pm 1.20}$}                 & \multicolumn{1}{c|}{$2.75^{\pm 3.03}$}                                              & \multicolumn{1}{c|}{$2.87^{\pm 1.81}$}                                              & \multicolumn{1}{c|}{\cellcolor[HTML]{C0C0C0}$1.22^{\pm 1.03}$} \\ \hline
\multicolumn{1}{|c|}{\textbf{Cancer}}                 & \multicolumn{1}{c|}{\cellcolor[HTML]{C0C0C0}$0.00^{\pm 0.00}$} & \multicolumn{1}{c|}{\cellcolor[HTML]{C0C0C0}$0.00^{\pm 0.00}$} & \multicolumn{1}{c|}{$0.83^{\pm 0.12}$}                   & \multicolumn{1}{c|}{\cellcolor[HTML]{C0C0C0}$0.00^{\pm 0.00}$} & \multicolumn{1}{c|}{\cellcolor[HTML]{C0C0C0}$0.00^{\pm 0.00}$} & \multicolumn{1}{c|}{$2.00^{\pm 0.86}$}                 & \multicolumn{1}{c|}{$0.16^{\pm 0.37}$}                                              & \multicolumn{1}{c|}{$3.53^{\pm 1.86}$}                                              & \multicolumn{1}{c|}{\cellcolor[HTML]{C0C0C0}$0.00^{\pm 0.00}$} \\ \hline
\multicolumn{1}{|c|}{\textbf{Climate}}                & \multicolumn{1}{c|}{$0.14^{\pm 0.35}$}                                              & \multicolumn{1}{c|}{$1.67^{\pm 3.13}$}                                              & \multicolumn{1}{c|}{$1.83^{\pm 0.17}$}                   & \multicolumn{1}{c|}{\cellcolor[HTML]{C0C0C0}$0.00^{\pm 0.00}$} & \multicolumn{1}{c|}{$0.83^{\pm 0.50}$}                                              & \multicolumn{1}{c|}{$4.42^{\pm 1.21}$}                 & \multicolumn{1}{c|}{$2.00^{\pm 0.00}$}                                              & \multicolumn{1}{c|}{$0.60^{\pm 0.80}$}                                              & \multicolumn{1}{c|}{$0.40^{\pm 0.80}$}                                              \\ \hline
\multicolumn{1}{|c|}{\textbf{Covid 1}}                & \multicolumn{1}{c|}{\cellcolor[HTML]{C0C0C0}$0.00^{\pm 0.00}$} & \multicolumn{1}{c|}{\cellcolor[HTML]{C0C0C0}$0.00^{\pm 0.00}$} & \multicolumn{1}{c|}{$0.25^{\pm 0.08}$}                   & \multicolumn{1}{c|}{\cellcolor[HTML]{C0C0C0}$0.00^{\pm 0.00}$} & \multicolumn{1}{c|}{$0.17^{\pm 0.14}$}                                              & \multicolumn{1}{c|}{$1.00^{\pm 0.74}$}                 & \multicolumn{1}{c|}{$1.00^{\pm 0.0}$}                                               & \multicolumn{1}{c|}{$0.55^{\pm 0.78}$}                                              & \multicolumn{1}{c|}{\cellcolor[HTML]{C0C0C0}$0.00^{\pm 0.00}$} \\ \hline
\multicolumn{1}{|c|}{\textbf{Covid 2}}                & \multicolumn{1}{c|}{$0.50^{\pm 0.50}$}                                              & \multicolumn{1}{c|}{$0.50^{\pm 0.50}$}                                              & \multicolumn{1}{c|}{$1.08^{\pm 0.49}$}                   & \multicolumn{1}{c|}{$0.50^{\pm 0.50}$}                                              & \multicolumn{1}{c|}{$0.50^{\pm 0.50}$}                                              & \multicolumn{1}{c|}{$1.67^{\pm 0.88}$}                 & \multicolumn{1}{c|}{$1.66^{\pm 0.47}$}                                              & \multicolumn{1}{c|}{\cellcolor[HTML]{C0C0C0}$0.00^{\pm 0.00}$} & \multicolumn{1}{c|}{\cellcolor[HTML]{C0C0C0}$0.00^{\pm 0.00}$} \\ \hline
\multicolumn{1}{|c|}{\textbf{Covid 3}}                & \multicolumn{1}{c|}{$0.50^{\pm 0.50}$}                                              & \multicolumn{1}{c|}{$1.62^{\pm 1.41}$}                                              & \multicolumn{1}{c|}{$0.75^{\pm 0.42}$}                   & \multicolumn{1}{c|}{$1.25^{\pm 0.42}$}                                              & \multicolumn{1}{c|}{$1.17^{\pm 0.75}$}                                              & \multicolumn{1}{c|}{$2.64^{\pm 1.25}$}                 & \multicolumn{1}{c|}{\cellcolor[HTML]{C0C0C0}$0.00^{\pm 0.00}$} & \multicolumn{1}{c|}{$1.00^{\pm 0.50}$}                                              & \multicolumn{1}{c|}{$1.00^{\pm 0.50}$}                                              \\ \hline
\multicolumn{1}{|c|}{\textbf{Covid 4}}                & \multicolumn{1}{c|}{$1.00^{\pm 0.73}$}                                              & \multicolumn{1}{c|}{$1.03^{\pm 0.72}$}                                              & \multicolumn{1}{c|}{$2.25^{\pm 0.67}$}                   & \multicolumn{1}{c|}{$1.00^{\pm 0.71}$}                                              & \multicolumn{1}{c|}{$1.00^{\pm 0.71}$}                                              & \multicolumn{1}{c|}{$3.49^{\pm 1.49}$}                 & \multicolumn{1}{c|}{$0.83^{\pm 0.68}$}                                              & \multicolumn{1}{c|}{\cellcolor[HTML]{C0C0C0}$0.00^{\pm 0.00}$} & \multicolumn{1}{c|}{\cellcolor[HTML]{C0C0C0}$0.00^{\pm 0.00}$} \\ \hline
\multicolumn{1}{|c|}{\textbf{Genetic}}                & \multicolumn{1}{c|}{\cellcolor[HTML]{C0C0C0}$0.00^{\pm 0.00}$} & \multicolumn{1}{c|}{$0.00^{\pm 0.00}$}                                              & \multicolumn{1}{c|}{$0.50^{\pm 0.00}$}                   & \multicolumn{1}{c|}{\cellcolor[HTML]{C0C0C0}$0.00^{\pm 0.00}$} & \multicolumn{1}{c|}{$0.25^{\pm 0.25}$}                                              & \multicolumn{1}{c|}{$2.75^{\pm 0.93}$}                 & \multicolumn{1}{c|}{\cellcolor[HTML]{C0C0C0}$0.00^{\pm 0.00}$} & \multicolumn{1}{c|}{$1.10^{\pm 0.70}$}                                              & \multicolumn{1}{c|}{$1.20^{\pm 0.40}$}                                              \\ \hline
\multicolumn{1}{|c|}{\textbf{MSU}}                    & \multicolumn{1}{c|}{$1.29^{\pm 0.70}$}                                              & \multicolumn{1}{c|}{$1.56^{\pm 1.07}$}                                              & \multicolumn{1}{c|}{$1.33^{\pm 0.17}$}                   & \multicolumn{1}{c|}{\cellcolor[HTML]{C0C0C0}$1.00^{\pm 0.00}$} & \multicolumn{1}{c|}{\cellcolor[HTML]{C0C0C0}$1.00^{\pm 0.00}$} & \multicolumn{1}{c|}{$3.00^{\pm 1.32}$}                 & \multicolumn{1}{c|}{$1.33^{\pm 0.74}$}                                              & \multicolumn{1}{c|}{$3.22^{\pm 1.73}$}                                              & \multicolumn{1}{c|}{$2.00^{\pm 0.58}$}                                              \\ \hline
\multicolumn{1}{|c|}{\textbf{Neighbor}}               & \multicolumn{1}{c|}{$5.65^{\pm 3.12}$}                                              & \multicolumn{1}{c|}{\cellcolor[HTML]{C0C0C0}$0.00^{\pm 0.00}$} & \multicolumn{1}{c|}{$2.33^{\pm 0.12}$}                   & \multicolumn{1}{c|}{$0.88^{\pm 0.14}$}                                              & \multicolumn{1}{c|}{$1.21^{\pm 0.14}$}                                              & \multicolumn{1}{c|}{$4.30^{\pm 1.59}$}                 & \multicolumn{1}{c|}{$3.00^{\pm 0.00}$}                                              & \multicolumn{1}{c|}{$5.96^{\pm 2.64}$}                                              & \multicolumn{1}{c|}{$4.89^{\pm 2.42}$}                                              \\ \hline
\multicolumn{1}{|c|}{\textbf{Sachs}}                  & \multicolumn{1}{c|}{$9.83^{\pm 3.17}$}                                              & \multicolumn{1}{c|}{$10.02^{\pm 3.12}$}                                             & \multicolumn{1}{c|}{$8.77^{\pm 1.73}$}                   & \multicolumn{1}{c|}{$7.38^{\pm 1.97}$}                                              & \multicolumn{1}{c|}{$8.42^{\pm 1.61}$}                                              & \multicolumn{1}{c|}{$9.31^{\pm 2.57}$}                 & \multicolumn{1}{c|}{$10.33^{\pm 9.19}$}                                             & \multicolumn{1}{c|}{$7.40^{\pm 2.04}$}                                              & \multicolumn{1}{c|}{\cellcolor[HTML]{C0C0C0}$5.50^{\pm 0.50}$} \\ \hline
\multicolumn{1}{|c|}{\textbf{Supermarket}}            & \multicolumn{1}{c|}{$6.38^{\pm 2.63}$}                                              & \multicolumn{1}{c|}{$9.11^{\pm 3.64}$}                                              & \multicolumn{1}{c|}{$4.31^{\pm 1.30}$}                   & \multicolumn{1}{c|}{$5.12^{\pm 1.19}$}                                              & \multicolumn{1}{c|}{$4.21^{\pm 0.88}$}                                              & \multicolumn{1}{c|}{$6.49^{\pm 1.83}$}                 & \multicolumn{1}{c|}{$5.0^{\pm 2.23}$}                                               & \multicolumn{1}{c|}{$6.45^{\pm 2.58}$}                                              & \multicolumn{1}{c|}{\cellcolor[HTML]{C0C0C0}$1.56^{\pm 1.26}$} \\ \hline
\multicolumn{1}{l}{}                                  & \multicolumn{1}{l}{}                                                                & \multicolumn{1}{l}{}                                                                & \multicolumn{1}{l}{}                                   &                                                                                     &                                                                                     &                                                        & \multicolumn{1}{l}{}                                                                & \multicolumn{1}{l}{}                                                                & \multicolumn{1}{l}{}                                                               
\end{tabular}
\caption{$\mathcal{D}_{top}$ ($\downarrow$) of the estimated causal orders. Best results are highlighted in gray.}
\label{tab:results}
\end{table}

\begin{table}
\centering
\scriptsize
\setlength{\tabcolsep}{2pt}
\begin{tabular}{cccccccccc}
\hline
\multicolumn{1}{|c|}{\multirow{2}{*}{\textbf{Graph}}} & \multicolumn{1}{c|}{\multirow{2}{*}{\textbf{PC}}}                                   & \multicolumn{1}{c|}{\multirow{2}{*}{\textbf{GES}}} & \multicolumn{1}{c|}{\multirow{2}{*}{\textbf{NOTEARS}}} & \multicolumn{1}{c|}{\multirow{2}{*}{\textbf{BOSS}}} & \multicolumn{1}{c|}{\multirow{2}{*}{\textbf{GRaSP}}} & \multicolumn{1}{c|}{\multirow{2}{*}{\textbf{GOBNILP}}} & \multicolumn{1}{c|}{\multirow{2}{*}{\textbf{Triplet}}}                              & \multicolumn{2}{c|}{\textbf{MATS (Our)}}                                                                                                                                  \\ \cline{9-10} 
\multicolumn{1}{|c|}{}                                & \multicolumn{1}{c|}{}                                                               & \multicolumn{1}{c|}{}                              & \multicolumn{1}{c|}{}                                  & \multicolumn{1}{c|}{}                               & \multicolumn{1}{c|}{}                                & \multicolumn{1}{c|}{}                                  & \multicolumn{1}{c|}{}                                                               & \multicolumn{1}{c|}{\textbf{MPDAG}}                                                 & \multicolumn{1}{c|}{\textbf{$\pi$}}                                                 \\ \hline
\multicolumn{1}{|c|}{\textbf{Asia}}                   & \multicolumn{1}{c|}{$8.37^{\pm2.27}$}                                               & \multicolumn{1}{c|}{$6.62^{\pm0.99}$}              & \multicolumn{1}{c|}{$6.06^{\pm0.36}$}                   & \multicolumn{1}{c|}{$7.44^{\pm0.29}$}               & \multicolumn{1}{c|}{$7.25^{\pm0.08}$}                & \multicolumn{1}{c|}{$6.94^{\pm0.37}$}                  & \multicolumn{1}{c|}{$2.75^{\pm 3.03}$}                                              & \multicolumn{1}{c|}{$2.87^{\pm 1.81}$}                                              & \multicolumn{1}{c|}{\cellcolor[HTML]{C0C0C0}$1.22^{\pm 1.03}$} \\ \hline
\multicolumn{1}{|c|}{\textbf{Cancer}}                 & \multicolumn{1}{c|}{$3.67^{\pm0.47}$}                                               & \multicolumn{1}{c|}{$3.00^{\pm0.00}$}              & \multicolumn{1}{c|}{$2.62^{\pm0.49}$}                   & \multicolumn{1}{c|}{$3.92^{\pm0.08}$}               & \multicolumn{1}{c|}{$3.92^{\pm0.08}$}                & \multicolumn{1}{c|}{$3.25^{\pm0.63}$}                  & \multicolumn{1}{c|}{$0.16^{\pm 0.37}$}                                              & \multicolumn{1}{c|}{$3.53^{\pm 1.86}$}                                              & \multicolumn{1}{c|}{\cellcolor[HTML]{C0C0C0}$0.00^{\pm 0.00}$} \\ \hline
\multicolumn{1}{|c|}{\textbf{Climate}}                & \multicolumn{1}{c|}{$8.00^{\pm2.52}$}                                               & \multicolumn{1}{c|}{$8.00^{\pm0.00}$}              & \multicolumn{1}{c|}{$5.98^{\pm0.43}$}                   & \multicolumn{1}{c|}{$7.92^{\pm0.08}$}               & \multicolumn{1}{c|}{$7.67^{\pm0.00}$}                & \multicolumn{1}{c|}{$8.00^{\pm0.00}$}                  & \multicolumn{1}{c|}{$2.00^{\pm 0.00}$}                                              & \multicolumn{1}{c|}{$0.60^{\pm 0.80}$}                                              & \multicolumn{1}{c|}{\cellcolor[HTML]{C0C0C0}$0.40^{\pm 0.80}$} \\ \hline
\multicolumn{1}{|c|}{\textbf{Covid 1}}                & \multicolumn{1}{c|}{\cellcolor[HTML]{C0C0C0} $0.00^{\pm0.00}$} & \multicolumn{1}{c|}{$1.00^{\pm0.00}$}              & \multicolumn{1}{c|}{$2.00^{\pm0.00}$}                   & \multicolumn{1}{c|}{$2.00^{\pm0.00}$}               & \multicolumn{1}{c|}{$2.00^{\pm0.00}$}                & \multicolumn{1}{c|}{$1.54^{\pm0.37}$}                  & \multicolumn{1}{c|}{$1.00^{\pm 0.0}$}                                               & \multicolumn{1}{c|}{$0.55^{\pm 0.78}$}                                              & \multicolumn{1}{c|}{\cellcolor[HTML]{C0C0C0}$0.00^{\pm 0.00}$} \\ \hline
\multicolumn{1}{|c|}{\textbf{Covid 2}}                & \multicolumn{1}{c|}{$1.50^{\pm0.50}$}                                               & \multicolumn{1}{c|}{$1.00^{\pm0.00}$}              & \multicolumn{1}{c|}{$2.75^{\pm0.20}$}                   & \multicolumn{1}{c|}{$3.00^{\pm0.00}$}               & \multicolumn{1}{c|}{$3.00^{\pm0.00}$}                & \multicolumn{1}{c|}{$2.25^{\pm0.59}$}                  & \multicolumn{1}{c|}{$1.66^{\pm 0.47}$}                                              & \multicolumn{1}{c|}{\cellcolor[HTML]{C0C0C0}$0.00^{\pm 0.00}$} & \multicolumn{1}{c|}{\cellcolor[HTML]{C0C0C0}$0.00^{\pm 0.00}$} \\ \hline
\multicolumn{1}{|c|}{\textbf{Covid 3}}                & \multicolumn{1}{c|}{$2.00^{\pm0.00}$}                                               & \multicolumn{1}{c|}{$2.00^{\pm0.00}$}              & \multicolumn{1}{c|}{$4.46^{\pm0.34}$}                   & \multicolumn{1}{c|}{$5.00^{\pm0.00}$}               & \multicolumn{1}{c|}{$5.00^{\pm0.00}$}                & \multicolumn{1}{c|}{$4.45^{\pm0.45}$}                  & \multicolumn{1}{c|}{\cellcolor[HTML]{C0C0C0}$0.00^{\pm 0.00}$} & \multicolumn{1}{c|}{$1.00^{\pm 0.50}$}                                              & \multicolumn{1}{c|}{$1.00^{\pm 0.50}$}                                              \\ \hline
\multicolumn{1}{|c|}{\textbf{Covid 4}}                & \multicolumn{1}{c|}{$3.17^{\pm1.67}$}                                               & \multicolumn{1}{c|}{$4.00^{\pm0.00}$}              & \multicolumn{1}{c|}{$3.82^{\pm0.31}$}                   & \multicolumn{1}{c|}{$5.50^{\pm0.14}$}               & \multicolumn{1}{c|}{$5.50^{\pm0.14}$}                & \multicolumn{1}{c|}{$5.67^{\pm0.33}$}                  & \multicolumn{1}{c|}{$0.83^{\pm 0.68}$}                                              & \multicolumn{1}{c|}{\cellcolor[HTML]{C0C0C0}$0.00^{\pm 0.00}$} & \multicolumn{1}{c|}{\cellcolor[HTML]{C0C0C0}$0.00^{\pm 0.00}$} \\ \hline
\multicolumn{1}{|c|}{\textbf{Genetic}}                & \multicolumn{1}{c|}{$4.00^{\pm0.00}$}                                               & \multicolumn{1}{c|}{$6.00^{\pm1.26}$}              & \multicolumn{1}{c|}{$2.90^{\pm0.73}$}                   & \multicolumn{1}{c|}{$4.92^{\pm0.08}$}               & \multicolumn{1}{c|}{$4.92^{\pm0.08}$}                & \multicolumn{1}{c|}{$3.65^{\pm0.73}$}                  & \multicolumn{1}{c|}{\cellcolor[HTML]{C0C0C0}$0.00^{\pm 0.00}$} & \multicolumn{1}{c|}{$1.10^{\pm 0.70}$}                                              & \multicolumn{1}{c|}{$1.20^{\pm 0.40}$}                                              \\ \hline
\multicolumn{1}{|c|}{\textbf{MSU}}                    & \multicolumn{1}{c|}{$1.50^{\pm0.50}$}                                               & \multicolumn{1}{c|}{$3.40^{\pm0.80}$}              & \multicolumn{1}{c|}{$4.74^{\pm0.71}$}                   & \multicolumn{1}{c|}{$5.92^{\pm0.08}$}               & \multicolumn{1}{c|}{$5.92^{\pm0.08}$}                & \multicolumn{1}{c|}{$4.22^{\pm0.91}$}                  & \multicolumn{1}{c|}{\cellcolor[HTML]{C0C0C0}$1.33^{\pm 0.74}$} & \multicolumn{1}{c|}{$3.22^{\pm 1.73}$}                                              & \multicolumn{1}{c|}{$2.00^{\pm 0.58}$}                                              \\ \hline
\multicolumn{1}{|c|}{\textbf{Neighbor}}               & \multicolumn{1}{c|}{$6.67^{\pm1.25}$}                                               & \multicolumn{1}{c|}{$7.20^{\pm1.60}$}              & \multicolumn{1}{c|}{$5.65^{\pm0.54}$}                   & \multicolumn{1}{c|}{$7.75^{\pm0.08}$}               & \multicolumn{1}{c|}{$7.75^{\pm0.08}$}                & \multicolumn{1}{c|}{$7.00^{\pm0.58}$}                  & \multicolumn{1}{c|}{\cellcolor[HTML]{C0C0C0}$3.00^{\pm 0.00}$} & \multicolumn{1}{c|}{$5.96^{\pm 2.64}$}                                              & \multicolumn{1}{c|}{$4.89^{\pm 2.42}$}                                              \\ \hline
\multicolumn{1}{|c|}{\textbf{Sachs}}                  & \multicolumn{1}{c|}{$7.75^{\pm1.20}$}                                               & \multicolumn{1}{c|}{$10.02^{\pm3.12}$}             & \multicolumn{1}{c|}{$13.50^{\pm0.99}$}                  & \multicolumn{1}{c|}{$15.15^{\pm0.61}$}              & \multicolumn{1}{c|}{$14.95^{\pm0.76}$}               & \multicolumn{1}{c|}{$15.00^{\pm1.08}$}                 & \multicolumn{1}{c|}{$10.33^{\pm 9.19}$}                                             & \multicolumn{1}{c|}{$7.40^{\pm 2.04}$}                                              & \multicolumn{1}{c|}{\cellcolor[HTML]{C0C0C0}$5.50^{\pm 0.50}$} \\ \hline
\multicolumn{1}{|c|}{\textbf{Supermarket}}            & \multicolumn{1}{c|}{$12.33^{\pm2.36}$}                                              & \multicolumn{1}{c|}{$12.50^{\pm0.50}$}             & \multicolumn{1}{c|}{$8.08^{\pm0.58}$}                   & \multicolumn{1}{c|}{$10.92^{\pm0.08}$}              & \multicolumn{1}{c|}{$10.83^{\pm0.17}$}               & \multicolumn{1}{c|}{$11.50^{\pm0.50}$}                 & \multicolumn{1}{c|}{$5.0^{\pm 2.23}$}                                               & \multicolumn{1}{c|}{$6.45^{\pm 2.58}$}                                              & \multicolumn{1}{c|}{\cellcolor[HTML]{C0C0C0}$1.56^{\pm 1.26}$} \\ \hline
\multicolumn{1}{l}{}                                  & \multicolumn{1}{l}{}                                                                & \multicolumn{1}{l}{}                               & \multicolumn{1}{l}{}                                   &                                                     &                                                      &                                                        & \multicolumn{1}{l}{}                                                                & \multicolumn{1}{l}{}                                                                & \multicolumn{1}{l}{}                                                               
\end{tabular}
\caption{$\mathcal{D}_{top}$ ($\downarrow$) of the estimated causal orders with non-linear synthetic data. Best results are highlighted in gray.}
\label{tab:results_nonlinear}
\end{table}

\paragraph{\bf Baselines.} Concerning LLM-aided approaches, we compare to a state-of-the-art method for recovering causal orders, proposed in~\cite{Vashishtha_2025}, which we refer to as the triplet method, or Triplet\footnote{We implemented this method based on the paper and discussions with the authors.}.
Moreover, we evaluate our approach against more traditional causal discovery methods. In particular, we consider the PC algorithm with Fisher’s conditional independence test in the linear case and KCI for the non-linear one, GES\footnote{https://github.com/py-why/causal-learn}, NOTEARS\footnote{https://github.com/xunzheng/notears}, GRaSP~\citep{Lam22}, BOSS~\citep{Bryan23}, and GOBNILP~\citep{Cussens20}, using $1000$ data samples.
Finally, we experimented with a hybrid approach that orients edges in the skeleton recovered by PC, using the orders obtained from MATS and Triplet \textemdash results in Appendix~\ref{appendix:additional}.

\paragraph{\bf Graphs \& Datasets} We tested MATS and the baselines on $12$ causal graphs. 
Among these, $3$ are well-known causal DAGs included in the \texttt{bnlearn} library, namely Asia~\citep{Lauritzen_2018}, Cancer~\citep{Korb_2004}, and Sachs~\citep{Sachs_2005}.
Additionally, we tested $9$ causal graphs sourced from scientific literature in the fields of epidemiology and public health.
These causal DAGs include Covid~1, Covid~2, Covid~3~\citep{Griffith_2020}, Covid~4~\citep{Glemain_2024}, Genetic~\citep{Palmer_2012}, MSU~\citep{Piccininni_2023}, Neighborhood~\citep{Basile_2009}, Climate~\citep{Guevara_2024}, and Supermarket~\citep{Basile_2012}.
The datasets used for data-driven causal discovery, i.e., for PC, GES, and NOTEARS, and total effect estimation, are generated based on the true causal DAG.
We rely on both linear and non-linear synthetic data, which is meaningful since text-driven methods do not require parametric assumptions \textemdash more details on the graphs and the synthetic datasets in Appendix~\ref{appendix:implementation} and \ref{appendix:graphs}. 

\paragraph{\bf Evaluation}
Traditional metrics fall short in evaluating the error over the estimated causal orders. 
Indeed, there can be multiple causal orders that are consistent with the same causal DAG (Section~\ref{sec:background}).
To this end, we rely on a metric proposed in~\cite{Rolland_2022} that measures how well an estimated ordering respects the true causal graph, $\mathcal{G}$.
A correct causal order respects the ordering constraints of $\mathcal{G}$, meaning that if $X_i \succ_{\mathcal{G}} X_j$, then $X_i$ must precede $X_j$ in the estimated order $\hat{\pi}$.
The metric is then defined as follows:
\[ \mathcal{D}_{top}(\hat{\pi}, \mathcal{G}) =  \sum_{(X_i, X_j) \in \mathcal{C}} \mathbf{1}[\hat{\pi}(X_i) > \hat{\pi}(X_j)] \quad \text{where} \quad \mathcal{C} = \{(X_i, X_j) | X_i \succ_{\mathcal{G}} X_j\} \]
where $\mathbf{1}$ is an indicator function measuring the number of violated constraints in $\mathcal{C}$.
We have that $\mathcal{D}_{top}(\hat{\pi}, \mathcal{G}) = 0$ when $\hat{\pi}$ is a correct causal order for $\mathcal{G}$.
Evaluating the estimated MCAT using more traditional metrics, such as SHD, does not capture the quality of the estimated orders, since the graph is dense. 
For completeness, we provide a comparison using SHD in Appendix~\ref{appendix:additional}.
Concerning causal effect estimation, we measure the absolute error between the estimated and true total effect, $\epsilon_{ATE}$.

\begin{table}
\centering
\scriptsize
\setlength{\tabcolsep}{2pt}
\begin{tabular}{cccccccccc}
\hline
\multicolumn{1}{|c|}{\multirow{2}{*}{\textbf{Graph}}} & \multicolumn{1}{c|}{\multirow{2}{*}{\textbf{PC}}}                                   & \multicolumn{1}{c|}{\multirow{2}{*}{\textbf{GES}}}                                  & \multicolumn{1}{c|}{\multirow{2}{*}{\textbf{NOTEARS}}} & \multicolumn{1}{c|}{\multirow{2}{*}{\textbf{BOSS}}} & \multicolumn{1}{c|}{\multirow{2}{*}{\textbf{GRaSP}}}                                & \multicolumn{1}{c|}{\multirow{2}{*}{\textbf{GOBNILP}}} & \multicolumn{1}{c|}{\multirow{2}{*}{\textbf{Triplet}}}                               & \multicolumn{2}{c|}{\textbf{MATS (Our)}}                                                                                                                                  \\ \cline{9-10} 
\multicolumn{1}{|c|}{}                                & \multicolumn{1}{c|}{}                                                               & \multicolumn{1}{c|}{}                                                               & \multicolumn{1}{c|}{}                                  & \multicolumn{1}{c|}{}                               & \multicolumn{1}{c|}{}                                                               & \multicolumn{1}{c|}{}                                  & \multicolumn{1}{c|}{}                                                                & \multicolumn{1}{c|}{\textbf{MPDAG}}                                                 & \multicolumn{1}{c|}{\textbf{Causal Orders}}                                         \\ \hline
\multicolumn{1}{|c|}{\textbf{Asia}}                   & \multicolumn{1}{c|}{$0.26^{\pm 0.18}$}                                              & \multicolumn{1}{c|}{$0.26^{\pm 0.18}$}                                              & \multicolumn{1}{c|}{$0.16^{\pm 0.13}$}                   & \multicolumn{1}{c|}{$0.16^{\pm 0.12}$}              & \multicolumn{1}{c|}{$0.17^{\pm 0.13}$}                                              & \multicolumn{1}{c|}{$0.19^{\pm 0.09}$}                 & \multicolumn{1}{c|}{\cellcolor[HTML]{C0C0C0} $0.00^{\pm 0.00}$} & \multicolumn{1}{c|}{-}                                                              & \multicolumn{1}{c|}{\cellcolor[HTML]{C0C0C0}$0.00^{\pm 0.00}$} \\ \hline
\multicolumn{1}{|c|}{\textbf{Cancer}}                 & \multicolumn{1}{c|}{$0.03^{\pm 0.01}$}                                              & \multicolumn{1}{c|}{$0.03^{\pm 0.01}$}                                              & \multicolumn{1}{c|}{$0.01^{\pm 0.00}$}                   & \multicolumn{1}{c|}{$0.01^{\pm 0.00}$}              & \multicolumn{1}{c|}{$0.01^{\pm 0.00}$}                                              & \multicolumn{1}{c|}{$0.14^{\pm 0.08}$}                 & \multicolumn{1}{c|}{\cellcolor[HTML]{C0C0C0}$0.00^{\pm 0.00}$}  & \multicolumn{1}{c|}{-}                                                              & \multicolumn{1}{c|}{\cellcolor[HTML]{C0C0C0}$0.00^{\pm 0.00}$} \\ \hline
\multicolumn{1}{|c|}{\textbf{Climate}}                & \multicolumn{1}{c|}{$0.03^{\pm 0.03}$}                                              & \multicolumn{1}{c|}{$0.03^{\pm 0.03}$}                                              & \multicolumn{1}{c|}{$0.03^{\pm 0.01}$}                 & \multicolumn{1}{c|}{$0.01^{\pm 0.00}$}              & \multicolumn{1}{c|}{$0.08^{\pm 0.06}$}                                              & \multicolumn{1}{c|}{$0.20^{\pm 0.11}$}                 & \multicolumn{1}{c|}{\cellcolor[HTML]{C0C0C0}$0.01^{\pm 0.00}$}  & \multicolumn{1}{c|}{-}                                                              & \multicolumn{1}{c|}{$0.14^{\pm 0.24}$}                                              \\ \hline
\multicolumn{1}{|c|}{\textbf{Covid 1}}                & \multicolumn{1}{c|}{$0.03^{\pm 0.01}$}                                              & \multicolumn{1}{c|}{$0.03^{\pm 0.01}$}                                              & \multicolumn{1}{c|}{$0.01^{\pm 0.00}$}                   & \multicolumn{1}{c|}{$0.01^{\pm 0.00}$}              & \multicolumn{1}{c|}{$0.01^{\pm 0.00}$}                                              & \multicolumn{1}{c|}{$0.01^{\pm 0.00}$}                 & \multicolumn{1}{c|}{$0.01^{\pm 0.00}$}                                               & \multicolumn{1}{c|}{\cellcolor[HTML]{C0C0C0}$0.00^{\pm 0.00}$} & \multicolumn{1}{c|}{\cellcolor[HTML]{C0C0C0}$0.00^{\pm 0.00}$} \\ \hline
\multicolumn{1}{|c|}{\textbf{Covid 2}}                & \multicolumn{1}{c|}{$0.52^{\pm 0.24}$}                                              & \multicolumn{1}{c|}{$0.52^{\pm 0.24}$}                                              & \multicolumn{1}{c|}{$0.01^{\pm 0.00}$}                   & \multicolumn{1}{c|}{$0.01^{\pm 0.00}$}              & \multicolumn{1}{c|}{$0.01^{\pm 0.00}$}                                              & \multicolumn{1}{c|}{$0.01^{\pm 0.00}$}                 & \multicolumn{1}{c|}{$0.01^{\pm 0.00}$}                                               & \multicolumn{1}{c|}{\cellcolor[HTML]{C0C0C0}$0.00^{\pm 0.00}$} & \multicolumn{1}{c|}{\cellcolor[HTML]{C0C0C0}$0.00^{\pm 0.00}$} \\ \hline
\multicolumn{1}{|c|}{\textbf{Covid 3}}                & \multicolumn{1}{c|}{$0.03^{\pm 0.02}$}                                              & \multicolumn{1}{c|}{$0.13^{\pm 0.23}$}                                              & \multicolumn{1}{c|}{$0.02^{\pm 0.01}$}                   & \multicolumn{1}{c|}{$0.01^{\pm 0.00}$}              & \multicolumn{1}{c|}{$0.08^{\pm 0.06}$}                                              & \multicolumn{1}{c|}{$0.17^{\pm 0.15}$}                 & \multicolumn{1}{c|}{\cellcolor[HTML]{C0C0C0}$0.00^{\pm 0.00}$}  & \multicolumn{1}{c|}{$0.03^{\pm 0.06}$}                                              & \multicolumn{1}{c|}{$0.03^{\pm 0.06}$}                                              \\ \hline
\multicolumn{1}{|c|}{\textbf{Covid 4}}                & \multicolumn{1}{c|}{$0.03^{\pm 0.01}$}                                              & \multicolumn{1}{c|}{$0.05^{\pm 0.04}$}                                              & \multicolumn{1}{c|}{$0.14^{\pm 0.02}$}                   & \multicolumn{1}{c|}{$0.07^{\pm 0.08}$}              & \multicolumn{1}{c|}{$0.07^{\pm 0.08}$}                                              & \multicolumn{1}{c|}{$0.12^{\pm 0.05}$}                 & \multicolumn{1}{c|}{$0.13^{\pm 0.15}$}                                               & \multicolumn{1}{c|}{$0.01^{\pm 0.01}$}                                              & \multicolumn{1}{c|}{\cellcolor[HTML]{C0C0C0}$0.00^{\pm 0.00}$} \\ \hline
\multicolumn{1}{|c|}{\textbf{Genetic}}                & \multicolumn{1}{c|}{$0.02^{\pm 0.01}$}                                              & \multicolumn{1}{c|}{$0.02^{\pm 0.01}$}                                              & \multicolumn{1}{c|}{$0.01^{\pm 0.00}$}                   & \multicolumn{1}{c|}{$0.01^{\pm 0.00}$}              & \multicolumn{1}{c|}{$0.01^{\pm 0.00}$}                                              & \multicolumn{1}{c|}{$0.01^{\pm 0.00}$}                 & \multicolumn{1}{c|}{\cellcolor[HTML]{C0C0C0}$0.00^{\pm 0.00}$}  & \multicolumn{1}{c|}{-}                                                              & \multicolumn{1}{c|}{\cellcolor[HTML]{C0C0C0}$0.00^{\pm 0.00}$} \\ \hline
\multicolumn{1}{|c|}{\textbf{MSU}}                    & \multicolumn{1}{c|}{\cellcolor[HTML]{C0C0C0}$0.19^{\pm 0.14}$} & \multicolumn{1}{c|}{\cellcolor[HTML]{C0C0C0}$0.19^{\pm 0.14}$} & \multicolumn{1}{c|}{$0.26^{\pm 0.00}$}                   & \multicolumn{1}{c|}{$0.26^{\pm 0.00}$}              & \multicolumn{1}{c|}{$0.26^{\pm 0.00}$}                                              & \multicolumn{1}{c|}{$0.41^{\pm 0.12}$}                 & \multicolumn{1}{c|}{$0.25^{\pm 0.14}$}                                               & \multicolumn{1}{c|}{-}                                                              & \multicolumn{1}{c|}{$0.25^{\pm 0.14}$}                                              \\ \hline
\multicolumn{1}{|c|}{\textbf{Neighbor}}               & \multicolumn{1}{c|}{$0.45^{\pm 0.37}$}                                              & \multicolumn{1}{c|}{$0.26^{\pm 0.17}$}                                              & \multicolumn{1}{c|}{$0.17^{\pm 0.00}$}                   & \multicolumn{1}{c|}{$0.18^{\pm 0.00}$}              & \multicolumn{1}{c|}{\cellcolor[HTML]{C0C0C0}$0.14^{\pm 0.00}$} & \multicolumn{1}{c|}{$0.25^{\pm 0.07}$}                 & \multicolumn{1}{c|}{$0.28^{\pm 0.17}$}                                               & \multicolumn{1}{c|}{-}                                                              & \multicolumn{1}{c|}{$0.16^{\pm 0.07}$}                                              \\ \hline
\multicolumn{1}{|c|}{\textbf{Sachs}}                  & \multicolumn{1}{c|}{$0.08^{\pm 0.08}$}                                              & \multicolumn{1}{c|}{$0.15^{\pm 0.10}$}                                              & \multicolumn{1}{c|}{$0.23^{\pm 0.14}$}                   & \multicolumn{1}{c|}{$0.18^{\pm 0.12}$}              & \multicolumn{1}{c|}{$0.21^{\pm 0.02}$}                                              & \multicolumn{1}{c|}{$0.18^{\pm 0.08}$}                 & \multicolumn{1}{c|}{\cellcolor[HTML]{C0C0C0}$0.04^{\pm 0.05}$}  & \multicolumn{1}{c|}{-}                                                              & \multicolumn{1}{c|}{$0.11^{\pm 0.07}$}                                              \\ \hline
\multicolumn{1}{|c|}{\textbf{Supermarket}}            & \multicolumn{1}{c|}{$0.40^{\pm 0.35}$}                                              & \multicolumn{1}{c|}{$0.47^{\pm 0.52}$}                                              & \multicolumn{1}{c|}{$0.43^{\pm 0.15}$}                   & \multicolumn{1}{c|}{$0.37^{\pm 0.07}$}              & \multicolumn{1}{c|}{$0.42^{\pm 0.09}$}                                              & \multicolumn{1}{c|}{$0.45^{\pm 0.14}$}                 & \multicolumn{1}{c|}{\cellcolor[HTML]{C0C0C0}$0.30^{\pm 0.31}$}  & \multicolumn{1}{c|}{-}                                                              & \multicolumn{1}{c|}{$0.46^{\pm 0.25}$}                                              \\ \hline
\multicolumn{1}{l}{}                                  & \multicolumn{1}{l}{}                                                                & \multicolumn{1}{l}{}                                                                & \multicolumn{1}{l}{}                                   & \multicolumn{1}{l}{}                                & \multicolumn{1}{l}{}                                                                & \multicolumn{1}{l}{}                                   & \multicolumn{1}{l}{}                                                                 & \multicolumn{1}{l}{}                                                                & \multicolumn{1}{l}{}                                                               
\end{tabular}
    \caption{$\epsilon_{ATE}$ ($\downarrow$) of total effect estimation. Best results are highlighted in gray. If non identifiable the value is set to '-'.}
    \label{tab:causal_effect}
\end{table}

\paragraph{\bf Results.} Table~\ref{tab:results}, \ref{tab:results_nonlinear}, and \ref{tab:causal_effect} report evaluation results for causal order recovery in the linear case, the non-linear case, and total effect estimation, respectively. Both tables show the mean and standard deviation over $5$ runs for each method, using different random seeds.
For methods that return abstractions (CPDAGs produced by PC, GES, BOSS, GRaSP, and GOBNILP; MPDAGs; or classes of causal orders) rather than a single causal order, the reported error is computed by averaging over all admissible orders across configurations.
In the case of NOTEARS, which returns a single DAG, we first recover the corresponding CPDAG and evaluate the error over that, so as not to penalize it for returning a single representative of its equivalence class \textemdash results relative to the recovered DAG are provided in Appendix~\ref{appendix:additional}.
Additionally, this evaluation may penalize methods that return large equivalence classes with high variance. To provide a more complete picture, additional intraclass estimation results are reported in Appendix~\ref{appendix:additional}.
The results in Table~\ref{tab:results}, for PC, GES, NOTEARS, BOSS, GRaSP, and GOBNILP, as well as for total effect estimation, are obtained using linearly generated data, a setting that is favorable to traditional causal discovery methods. 
Indeed, as shown in Table~\ref{tab:results_nonlinear}, their performance deteriorates substantially when data are generated from a non-linear model.
In contrast, text-based methods such as MATS and Triplet do not rely on parametric assumptions.
In the linear setting, MATS achieves the lowest $\mathcal{D}_{top}$ error in $7$ out of the $12$ graphs, while in the non-linear setting it outperforms all other methods in $8$ out of the $12$ graphs. 
Moreover, in $4$ graphs, MATS recovers a class containing exclusively correct causal orders, both in the linear and non-linear case; in two additional cases, the resulting MPDAG is an acyclic tournament and therefore requires no further edge orientation.
Total effects are estimated using the adjustment formula in~\eqref{formula:generalized_adjustment}, implemented via linear regression.
Adjustment sets for data-driven methods are identified on the estimated graph using the generalized backdoor criterion (Definition~\ref{def:generlized_bd}).
The resulting estimation errors are reported in Table~\ref{tab:causal_effect}. Overall, we observe a clear correlation between the topological distance $\mathcal{D}_{top}$ and the absolute error in total effect estimation, $\epsilon_{ATE}$. 
Finally, we note that in the majority of cases, the estimated MPDAG does not permit identification of the total effect.

\section{Discussion \& Conclusion}
\label{sec:discussion}

The method described in this paper offers an approach leveraging LLMs as knowledge bases for retrieving abstractions of causal orders. 
This approach relies solely on textual descriptions of the variables and does not require faithfulness, parametric assumptions, or observational data.
A key insight behind the method is that natural language often leaves causal mechanisms implicit.
This implies that some causal relationships may not be directly stated in the text, yet the causal order remains intact.
We compared our methods with traditional approaches, such as PC, GES, NOTEARS, BOSS, GRaSP, and GOBNILP, as well as an LLM-aided method recovering causal order~\citep{Vashishtha_2025}. 
The results show that our approach can provide an accurate estimation most of the time, outperforming other methods.


\paragraph{\bf Background Knowledge.}
Note that the proposed method is compatible with the use of background knowledge \textemdash for instance, orientations provided by human experts or temporal priority. 
In particular, we can force orderings among variables by assigning specific values in the consistency matrix. 
For instance, if we know that $X_i$ is a predecessor of $X_j$: we can set $C_{X_i \succ X_j} = 1$, being maximally consistent, while $C_{X_j \succ X_i} = -\infty$, effectively removing the edge $X_j \to X_i$ from every maximally consistent graph.

\paragraph{Limitations.} The MATS algorithm has several limitations that should be taken into account.
Indeed, the accuracy of the estimation strongly relies on the consistency of the LLM. 
In the presence of an inconsistent knowledge base, we cannot guarantee the correctness of the class of orders retrieved.
Moreover, the computational complexity of the method can increase significantly with larger graphs.
This is primarily due to the computation of the consistency matrix, which has a quadratic complexity in relation to the number of nodes, and the ExactFAS algorithm, which is an NP-hard problem.
Note, however, that MATS can be made more efficient by introducing heuristic solvers for the FAS problem. 

\paragraph{Future Work.} Extensions of this work will focus on reducing computational complexity and enhancing the reliability of the LLM. 
The use of chain-of-thought based LLMs~\citep{Wei_2023} could potentially improve the accuracy of the method.
Moreover, in scenarios where a collection of documents relevant to a specific domain is available, we can enhance the reliability of the knowledge base by using Retrieval-Augmented Generation (RAG), proposed by~\cite{Lewis_2020}. 
This approach grounds replies generated by the LLM on a specific corpus of documents.


\section*{Acknowledgments}
This work was supported by the CIPHOD project (ANR-23-CPJ1-0212-01). 

\bibliographystyle{ACM-Reference-Format}
\bibliography{yourbibfile}

@article{Assaad_2022,
  author = {Assaad, Charles K. and Devijver, Emilie and Gaussier, Eric},
  title={Survey and Evaluation of Causal Discovery Methods for Time Series},
  year={2022},
  cdate={1640995200000},
  journal={J. Artif. Intell. Res.},
  volume={73},
  pages={767-819},
}

@article{Zheng_2024,
  author={Danna Zheng and Mirella Lapata and Jeff Z. Pan},
  title={Large Language Models as Reliable Knowledge Bases?},
  year={2024},
  cdate={1704067200000},
  journal={CoRR},
  volume={abs/2407.13578},
}

@inproceedings{Long_2023,
title={Causal Discovery with Language Models as Imperfect Experts},
author={Stephanie Long and Alexandre Pich{\'e} and Valentina Zantedeschi and Tibor Schuster and Alexandre Drouin},
booktitle={ICML 2023 Workshop on Structured Probabilistic Inference {\&} Generative Modeling},
year={2023},
}

@misc{Kadavath_2022,
      title={Language Models (Mostly) Know What They Know}, 
      author={Saurav Kadavath and Tom Conerly and Amanda Askell and Tom Henighan and Dawn Drain and Ethan Perez and Nicholas Schiefer and Zac Hatfield-Dodds and Nova DasSarma and Eli Tran-Johnson and Scott Johnston and Sheer El-Showk and Andy Jones and Nelson Elhage and Tristan Hume and Anna Chen and Yuntao Bai and Sam Bowman and Stanislav Fort and Deep Ganguli and Danny Hernandez and Josh Jacobson and Jackson Kernion and Shauna Kravec and Liane Lovitt and Kamal Ndousse and Catherine Olsson and Sam Ringer and Dario Amodei and Tom Brown and Jack Clark and Nicholas Joseph and Ben Mann and Sam McCandlish and Chris Olah and Jared Kaplan},
      year={2022},
      eprint={2207.05221},
      archivePrefix={arXiv},
      primaryClass={cs.CL},
}

@article{Long_2023a,
  title={Can large language models build causal graphs?},
  author={Long, Stephanie and Schuster, Tibor and Pich{\'e}, Alexandre},
  journal={arXiv preprint arXiv:2303.05279},
  year={2023}
}

@inproceedings{Cohrs_2023,
title={Large Language Models for Constrained-Based Causal Discovery},
author={Kai-Hendrik Cohrs and Emiliano Diaz and Vasileios Sitokonstantinou and Gherardo Varando and Gustau Camps-Valls},
booktitle={AAAI 2024 Workshop on ''Are Large Language Models Simply Causal Parrots?''},
year={2023},
}

@inproceedings{Vashishtha_2025,
title={Causal Order: The Key to Leveraging Imperfect Experts in Causal Inference},
author={Aniket Vashishtha and Abbavaram Gowtham Reddy and Abhinav Kumar and Saketh Bachu and Vineeth N. Balasubramanian and Amit Sharma},
booktitle={The Thirteenth International Conference on Learning Representations},
year={2025},
}

@article{Li_2022,
author = {Li, Dan-dan and Yang, Yang and Gao, Zi-yi and Zhao, Li-hua and Yang, Xue and Xu, Feng and Yu, Chao and Zhang, Xiu-lin and Wang, Xue-Qin and Wang, Li-hua and Su, Jian-Bin},
year = {2022},
month = {01},
pages = {},
title = {Sedentary lifestyle and body composition in type 2 diabetes},
volume = {14},
journal = {Diabetology \& Metabolic Syndrome},
doi = {10.1186/s13098-021-00778-6}
}

@article{Piccininni_2023,
  title={The effect of mobile stroke unit care on functional outcomes: an application of the front-door formula},
  author={Piccininni, Marco and Kurth, Tobias and Audebert, Heinrich J and Rohmann, Jessica L},
  journal={Epidemiology},
  volume={34},
  number={5},
  pages={712--720},
  year={2023},
  publisher={LWW}
}

@inproceedings{Wei_2023,
title={Chain of Thought Prompting Elicits Reasoning in Large Language Models},
author={Jason Wei and Xuezhi Wang and Dale Schuurmans and Maarten Bosma and brian ichter and Fei Xia and Ed H. Chi and Quoc V Le and Denny Zhou},
booktitle={Advances in Neural Information Processing Systems},
editor={Alice H. Oh and Alekh Agarwal and Danielle Belgrave and Kyunghyun Cho},
year={2022},
}

@article{Zecevic_2023,
title={Causal Parrots: Large Language Models May Talk Causality But Are Not Causal},
author={Matej Ze{\v{c}}evi{\'c} and Moritz Willig and Devendra Singh Dhami and Kristian Kersting},
journal={Transactions on Machine Learning Research},
issn={2835-8856},
year={2023},
note={}
}

@Article{Glemain_2024,
author={Glemain, Benjamin
and Assaad, Charles
and Ghosn, Walid
and Moulaire, Paul
and de Lamballerie, Xavier
and Zins, Marie
and Severi, Gianluca
and Touvier, Mathilde
and Deleuze, Jean-Fran{\c{c}}ois
and Lapidus, Nathana{\"e}l
and Carrat, Fabrice
and Ancel, Pierre-Yves
and Charles, Marie-Aline
and Kab, Sofiane
and Renuy, Adeline
and Le-Got, Stephane
and Ribet, Celine
and Pellicer, Mireille
and Wiernik, Emmanuel
and Goldberg, Marcel
and Artaud, Fanny
and Gerbouin-R{\'e}rolle, Pascale
and Enguix, M{\'e}lody
and Laplanche, Camille
and Gomes-Rima, Roselyn
and Hoang, Lyan
and Correia, Emmanuelle
and Barry, Alpha Amadou
and Senina, Nad{\`e}ge
and Allegre, Julien
and Szabo de Edelenyi, Fabien
and Druesne-Pecollo, Nathalie
and Esseddik, Younes
and Hercberg, Serge
and Deschasaux, M{\'e}lanie
and Benhammou, Val{\'e}rie
and Ritmi, Anass
and Marchand, Laetitia
and Zaros, Cecile
and Lordmi, Elodie
and Candea, Adriana
and de Visme, Sophie
and Simeon, Thierry
and Thierry, Xavier
and Geay, Bertrand
and Dufourg, Marie-Noelle
and Milcent, Karen
and Rahib, Delphine
and Lydie, Nathalie
and Lusivika-Nzinga, Clovis
and Pannetier, Gregory
and Lapidus, Nathanael
and Goderel, Isabelle
and Dorival, C{\'e}line
and Nicol, J{\'e}r{\^o}me
and Robineau, Olivier
and Lai, Cindy
and Belhadji, Liza
and Esperou, H{\'e}l{\`e}ne
and Couffin-Cadiergues, Sandrine
and Gagliolo, Jean-Marie
and Blanch{\'e}, H{\'e}l{\`e}ne
and S{\'e}baoun, Jean-Marc
and Beaudoin, Jean-Christophe
and Gressin, Laetitia
and Morel, Val{\'e}rie
and Ouili, Ouissam
and Ninove, Laetitia
and Priet, St{\'e}phane
and Villarroel, Paola Mariela Saba
and Fouri{\'e}, Toscane
and Mohamed Ali, Souand
and Amroun, Abdenour
and Seston, Morgan
and Ayhan, Nazli
and Pastorino, Boris
and group, SAPRIS-SERO study},
title={Revisiting the link between COVID-19 incidence and infection fatality rate during the first pandemic wave},
journal={Scientific Reports},
year={2025},
month={May},
day={05},
volume={15},
number={1},
pages={15638},
issn={2045-2322},
doi={10.1038/s41598-025-99078-6},
}

@article{Basile_2009,
author = {Chaix, Basile and Leal, Cinira and Evans, David},
year = {2009},
month = {11},
pages = {124-7},
title = {Neighborhood-level Confounding in Epidemiologic Studies Unavoidable Challenges, Uncertain Solutions},
volume = {21},
journal = {Epidemiology (Cambridge, Mass.)},
doi = {10.1097/EDE.0b013e3181c04e70}
}

@article{Basile_2012,
author = {Chaix, Basile and Bean, Kathy and Daniel, Mark and Zenk, Shannon and Kestens, Yan and Charreire, Hélène and Leal, Cinira and Thomas, Frédérique and Karusisi, Noëlla and Weber, Christiane and Oppert, Jean-Michel and Simon, Chantal and Merlo, Juan and Pannier, Bruce},
year = {2012},
month = {04},
pages = {e32908},
title = {Associations of Supermarket Characteristics with Weight Status and Body Fat: A Multilevel Analysis of Individuals within Supermarkets (RECORD Study)},
volume = {7},
journal = {PloS one},
doi = {10.1371/journal.pone.0032908}
}

@article{Griffith_2020,
  title={Collider bias undermines our understanding of COVID-19 disease risk and severity},
  author={Gareth J. Griffith and Tim T. Morris and Matthew J Tudball and Annie Herbert and Giulia Mancano and Lindsey Pike and Gemma C. Sharp and Jonathan A. C. Sterne and Tom M. Palmer and George Davey Smith and Kate Tilling and Luisa Zuccolo and Neil Martin Davies and Gibran Hemani},
  journal={Nature Communications},
  year={2020},
  volume={11},
}

@article{Palmer_2012,
  title={Using multiple genetic variants as instrumental variables for modifiable risk factors},
  author={Tom M. Palmer and Deborah A. Lawlor and Roger M. Harbord and Nuala A. Sheehan and Jonathan H. Tobias and Nicholas John Timpson and George Davey Smith and Jonathan A. C. Sterne},
  journal={Statistical Methods in Medical Research},
  year={2012},
  volume={21},
  pages={223 - 242},
}

@article{Barthelemy_1995,
title = {The reversing number of a diagraph},
journal = {Discrete Applied Mathematics},
volume = {60},
number = {1},
pages = {39-76},
year = {1995},
issn = {0166-218X},
doi = {https://doi.org/10.1016/0166-218X(94)00042-C},
author = {Jean-Pierre Barthélemy and Olivier Hudry and Garth Isaak and Fred S. Roberts and Barry Tesman}
}

@article{Korb_2004,
  title={Bayesian Artificial Intelligence},
  author={Kevin B. Korb and Ann E. Nicholson},
  booktitle={Computer science and data analysis series},
  year={2004},
}

@article{Sachs_2005,
  title={Causal Protein-Signaling Networks Derived from Multiparameter Single-Cell Data},
  author={Karen Sachs and Omar D. Perez and Dana Pe’er and Douglas A. Lauffenburger and Garry P. Nolan},
  journal={Science},
  year={2005},
  volume={308},
  pages={523 - 529},
}

@article{Lauritzen_2018,
    author = {Lauritzen, S. L. and Spiegelhalter, D. J.},
    title = {Local Computations with Probabilities on Graphical Structures and Their Application to Expert Systems},
    journal = {Journal of the Royal Statistical Society: Series B (Methodological)},
    volume = {50},
    number = {2},
    pages = {157-194},
    year = {2018},
    month = {12},
    issn = {0035-9246},
    doi = {10.1111/j.2517-6161.1988.tb01721.x},
    eprint = {https://academic.oup.com/jrsssb/article-pdf/50/2/157/49097926/jrsssb\_50\_2\_157.pdf},
}

@Article{Guevara_2024,
author={Barrero Guevara, Laura Andrea
and Kramer, Sarah C.
and Kurth, Tobias
and Domenech de Cell{\`e}s, Matthieu},
title={Causal inference concepts can guide research into the effects of climate on infectious diseases},
journal={Nature Ecology {\&} Evolution},
year={2025},
month={Feb},
day={01},
volume={9},
number={2},
pages={349-363},
issn={2397-334X},
doi={10.1038/s41559-024-02594-3},
}

@inproceedings{Si_2025,
title={Can {LLM}s Generate Novel Research Ideas? A Large-Scale Human Study with 100+ {NLP} Researchers},
author={Chenglei Si and Diyi Yang and Tatsunori Hashimoto},
booktitle={The Thirteenth International Conference on Learning Representations},
year={2025},
}

@book{Spirtes_2001,
  title={Causation, prediction, and search},
  author={Spirtes, Peter and Glymour, Clark and Scheines, Richard},
  year={2001},
  publisher={MIT press}
}

@book{Peters_2017,
author = {Peters, Jonas and Janzing, Dominik and Schlkopf, Bernhard},
title = {Elements of Causal Inference: Foundations and Learning Algorithms},
year = {2017},
isbn = {0262037319},
publisher = {The MIT Press},
}

@article{Glymour_2019,
AUTHOR={Glymour, Clark  and Zhang, Kun  and Spirtes, Peter },
TITLE={Review of Causal Discovery Methods Based on Graphical Models},
JOURNAL={Frontiers in Genetics},
VOLUME={10},
YEAR={2019},
DOI={10.3389/fgene.2019.00524},
ISSN={1664-8021},
}

@book{Pearl_2009,
  author    = {Pearl, Judea},
  publisher = {Cambridge University Press},
  title     = {Causality: Models, Reasoning and Inference},
  year      = {2009},
  address   = {Cambridge},
  edition   = {2},
  doi       = {https://doi.org/10.1017/CBO9780511803161}
}

@inproceedings{Jiralerspong_2024,
title={Efficient Causal Graph Discovery Using Large Language Models},
author={Thomas Jiralerspong and Xiaoyin Chen and Yash More and Vedant Shah and Yoshua Bengio},
booktitle={ICLR 2024 Workshop: How Far Are We From AGI},
year={2024},
}

@article{Kiciman_2024,
  title={Causal Reasoning and Large Language Models: Opening a New Frontier for Causality},
  author={Emre Kiciman and Robert Ness and Amit Sharma and Chenhao Tan},
  journal={Transactions on Machine Learning Research},
  issn={2835-8856},
  year={2024},
  note={Featured Certification}
}

@article{Cohrs_2025,
	title = {Large language models for causal hypothesis generation in science},
	volume = {6},
	doi = {10.1088/2632-2153/ada47f},
	number = {1},
	journal = {Machine Learning: Science and Technology},
	author = {Cohrs, Kai-Hendrik and Diaz, Emiliano and Sitokonstantinou, Vasileios and Varando, Gherardo and Camps-Valls, Gustau},
	month = jan,
	year = {2025},
	note = {Publisher: IOP Publishing},
	pages = {013001},
}

@article{Rolland_2022,
      title = 	 {Score Matching Enables Causal Discovery of Nonlinear Additive Noise Models},
  author =       {Rolland, Paul and Cevher, Volkan and Kleindessner, Matth{\"a}us and Russell, Chris and Janzing, Dominik and Sch{\"o}lkopf, Bernhard and Locatello, Francesco},
  booktitle = 	 {Proceedings of the 39th International Conference on Machine Learning},
  pages = 	 {18741--18753},
  year = 	 {2022},
  editor = 	 {Chaudhuri, Kamalika and Jegelka, Stefanie and Song, Le and Szepesvari, Csaba and Niu, Gang and Sabato, Sivan},
  volume = 	 {162},
  series = 	 {Proceedings of Machine Learning Research},
  month = 	 {17--23 Jul},
  publisher =    {PMLR}
}

@article{Maathuis_2015,
      author = {Marloes H. Maathuis and Diego Colombo},
      title = {{A generalized back-door criterion}},
      volume = {43},
      journal = {The Annals of Statistics},
      number = {3},
      publisher = {Institute of Mathematical Statistics},
      pages = {1060 -- 1088},
      year = {2015},
      doi = {10.1214/14-AOS1295},
}

@article{Perkovic_2020,
  title = 	 {Identifying causal effects in maximally oriented partially directed acyclic graphs},
  author =       {Perkovic, Emilija},
  booktitle = 	 {Proceedings of the 36th Conference on Uncertainty in Artificial Intelligence (UAI)},
  pages = 	 {530--539},
  year = 	 {2020},
  editor = 	 {Peters, Jonas and Sontag, David},
  volume = 	 {124},
  series = 	 {Proceedings of Machine Learning Research},
  month = 	 {03--06 Aug},
  publisher =    {PMLR},
}

@article{Perkovic_2017,
author = {Perkovic, Emilija and Textor, Johannes and Kalisch, Markus and Maathuis, Marloes H.},
title = {Complete graphical characterization and construction of adjustment sets in Markov equivalence classes of ancestral graphs},
year = {2017},
issue_date = {January 2017},
publisher = {JMLR.org},
volume = {18},
number = {1},
issn = {1532-4435},
journal = {J. Mach. Learn. Res.},
month = jan,
pages = {8132–8193},
numpages = {62}
}

@inproceedings{Manakul_2023,
    title = "{S}elf{C}heck{GPT}: Zero-Resource Black-Box Hallucination Detection for Generative Large Language Models",
    author = "Manakul, Potsawee  and
      Liusie, Adian  and
      Gales, Mark",
    editor = "Bouamor, Houda  and
      Pino, Juan  and
      Bali, Kalika",
    booktitle = "Proceedings of the 2023 Conference on Empirical Methods in Natural Language Processing",
    month = dec,
    year = "2023",
    address = "Singapore",
    publisher = "Association for Computational Linguistics",
    doi = "10.18653/v1/2023.emnlp-main.557",
    pages = "9004--9017"
}

@article{Savage_2024,
  title={Large language model uncertainty measurement and calibration for medical diagnosis and treatment},
  author={Savage, Thomas and Wang, John and Gallo, Robert and Boukil, Abdessalem and Patel, Vishwesh and Ahmad Safavi-Naini, Seyed Amir and Soroush, Ali and Chen, Jonathan H},
  journal={medRxiv},
  pages={2024--06},
  year={2024},
  publisher={Cold Spring Harbor Laboratory Press}
}

@article{Perkovic_2017b,
  title={Interpreting and using CPDAGs with background knowledge},
  author={Perkovi{\'c}, Emilija and Kalisch, Markus and Maathuis, Maloes H},
  journal={Association for Uncertainty in Artificial Intelligence (UAI)},
  year={2017}
}

@inproceedings{Meek_1995,
author = {Meek, Christopher},
title = {Causal inference and causal explanation with background knowledge},
year = {1995},
isbn = {1558603859},
publisher = {Morgan Kaufmann Publishers Inc.},
address = {San Francisco, CA, USA},
booktitle = {Proceedings of the Eleventh Conference on Uncertainty in Artificial Intelligence},
pages = {403–410},
numpages = {8},
location = {Montr\'{e}al, Qu\'{e}, Canada},
series = {UAI'95}
}

@misc{Venkateswaran_2024,
      title={Towards Complete Causal Explanation with Expert Knowledge}, 
      author={Aparajithan Venkateswaran and Emilija Perković},
      year={2024},
      eprint={2407.07338},
      archivePrefix={arXiv},
      primaryClass={stat.ML},
}

@inproceedings{Lewis_2020,
 author = {Lewis, Patrick and Perez, Ethan and Piktus, Aleksandra and Petroni, Fabio and Karpukhin, Vladimir and Goyal, Naman and K\"{u}ttler, Heinrich and Lewis, Mike and Yih, Wen-tau and Rockt\"{a}schel, Tim and Riedel, Sebastian and Kiela, Douwe},
 booktitle = {Advances in Neural Information Processing Systems},
 editor = {H. Larochelle and M. Ranzato and R. Hadsell and M.F. Balcan and H. Lin},
 pages = {9459--9474},
 publisher = {Curran Associates, Inc.},
 title = {Retrieval-Augmented Generation for Knowledge-Intensive NLP Tasks},
 volume = {33},
 year = {2020}
}

@article{Baharev_2021,
author = {Baharev, Ali and Schichl, Hermann and Neumaier, Arnold and Achterberg, Tobias},
title = {An Exact Method for the Minimum Feedback Arc Set Problem},
year = {2021},
issue_date = {December 2021},
publisher = {Association for Computing Machinery},
address = {New York, NY, USA},
volume = {26},
issn = {1084-6654},
doi = {10.1145/3446429},
journal = {ACM J. Exp. Algorithmics},
month = apr,
articleno = {1.4},
numpages = {28}
}

@article{Eades_1993,
author = {Eades, Peter and Lin, Xuemin and Smyth, W. F.},
title = {A fast and effective heuristic for the feedback arc set problem},
year = {1993},
issue_date = {Oct. 18, 1993},
publisher = {Elsevier North-Holland, Inc.},
address = {USA},
volume = {47},
number = {6},
issn = {0020-0190},
doi = {10.1016/0020-0190(93)90079-O},
journal = {Inf. Process. Lett.},
month = oct,
pages = {319–323},
numpages = {5}
}

@incollection{Karp_1972,
  title={Reducibility among combinatorial problems},
  author={Karp, Richard M},
  booktitle={50 Years of Integer Programming 1958-2008: from the Early Years to the State-of-the-Art},
  pages={219--241},
  year={1972},
  publisher={Springer}
}

@InProceedings{Lam22,
  title = 	 {Greedy relaxations of the sparsest permutation algorithm},
  author =       {Lam, Wai-Yin and Andrews, Bryan and Ramsey, Joseph},
  booktitle = 	 {Proceedings of the Thirty-Eighth Conference on Uncertainty in Artificial Intelligence},
  pages = 	 {1052--1062},
  year = 	 {2022},
  editor = 	 {Cussens, James and Zhang, Kun},
  volume = 	 {180},
  series = 	 {Proceedings of Machine Learning Research},
  month = 	 {01--05 Aug},
  publisher =    {PMLR}
}

@inproceedings{Bryan23,
 author = {Andrews, Bryan and Ramsey, Joseph and Sanchez Romero, Ruben and Camchong, Jazmin and Kummerfeld, Erich},
 booktitle = {Advances in Neural Information Processing Systems},
 editor = {A. Oh and T. Naumann and A. Globerson and K. Saenko and M. Hardt and S. Levine},
 pages = {63945--63956},
 publisher = {Curran Associates, Inc.},
 title = {Fast Scalable and Accurate Discovery of DAGs Using the Best Order Score Search and Grow Shrink Trees},
 volume = {36},
 year = {2023}
}

@InProceedings{Cussens20,
  title = 	 {GOBNILP: Learning Bayesian network structure with integer programming},
  author =       {Cussens, James},
  booktitle = 	 {Proceedings of the 10th International Conference on Probabilistic Graphical Models},
  pages = 	 {605--608},
  year = 	 {2020},
  editor = 	 {Jaeger, Manfred and Nielsen, Thomas Dyhre},
  volume = 	 {138},
  series = 	 {Proceedings of Machine Learning Research},
  month = 	 {23--25 Sep},
  publisher =    {PMLR}
}

\appendix
\section{Appendix}

\section*{Appendix}

The appendix is organized as follows:
\begin{itemize}
    \item Appendix~\ref{appendix:proofs}, contains the proofs of the propositions and theorems presented in the main text.
    \item Appendix~\ref{appendix:implementation}, contains details regarding the implementation of the MATS algorithm.
    \item Appendix~\ref{appendix:additional}, contains additional results.
    \item Appendix~\ref{appendix:graphs}, contains details on the graphs used in the experiments and the prompts used to query the LLM.
\end{itemize}

\section{Proofs}
\label{appendix:proofs}

\begin{proof}[Proposition~\ref{prop:1}]
    Given a DAG $\mathcal{G} = (\mathbb{V}, \mathbb{E})$, a consistent knowledge base induces a consistency matrix $W$ such that:
    for every pair of variables $X_i$ and $X_j$, if $X_i \succ_{\mathcal{G}} X_j$ then $C_{X_i \succ X_j} \ge C_{X_j \succ X_i}$.
    If we build a semi-complete PDG $\mathcal{S}$ from the consistency matrix $W$, it follows that:
    \begin{itemize}
        \item If $C_{X_i \succ X_j} > C_{X_j \succ X_i}$, then we orient the edge as $X_i \rightarrow X_j$; the ensemble of edges oriented in this way does not contain directed cycles since it is a directed reflection of the DAG $\mathcal{G}$.
        \item If $C_{X_i \succ X_j} = C_{X_j \succ X_i}$, then we orient the edge as $X_i - X_j$; this does not introduce any directed cycle.
    \end{itemize}
\end{proof}

\begin{proof}[Proposition~\ref{prop:2}]
    The proof follows from Proposition~\ref{prop:1}. 
    If the expert is consistent, the maximally consistent semi-complete PDG $\mathcal{S}$ does not contain directed cycles, meaning that it is a dense Partially Directed Acyclic Graph (PDAG).
    Additionally, since the graph is fully connected, there cannot be unshielded colliders.
    Thus, we can use R2 exclusively to maximize the orientation of the PDAG.
\end{proof}

\begin{proof}[Proposition~\ref{prop:3}]
    Substituting directed edges in a cycle with undirected ones transforms the Semi-Complete PDG into a dense PDAG.
    Applying the Meek rules, in particular R2, to the PDAG, we get a maximally oriented PDAG (Proposition~\ref{prop:2}).
\end{proof}

\begin{proof}[Proposition~\ref{prop:4}]
    Since the MPDAG is dense, between every pair of variables there is an edge, either directed or undirected.
    If the treatment and the outcome are connected by an undirected edge, it means that the total effect is not identifiable by adjustment.
    If the treatment and the outcome are connected by a directed edge, we can reduce the problem of identifiability to triplets of variables. 
    If any of the subgraphs containing the treatment, the outcome, and any other variables are of the form in Figure~\ref{fig:non_identifiable_subgraphs}, we can see that we cannot distinguish between a collider and a non-collider.
    Thus, the total effect is not identifiable.
\end{proof}

\begin{proof}[Corollary~\ref{corollary:1}]
    The proof follows from Proposition~\ref{prop:4}.
\end{proof}

\begin{proof}[Theorem~\ref{thm:1}]
We define the following notation in reference to Algorithm~\ref{alg:mats}:
\begin{itemize}
    \item $\texttt{maxScore}$ is the maximal consistency score; thus, given a maximally consistent acyclic tournament $\mathcal{T} = (\mathbb{V}, \mathbb{E}_{\mathcal{T}})$:
    $$\texttt{maxScore} = \sum_{(i, j) \in \mathbb{E}_{\mathcal{T}}} W[i, j]$$
    \item The method \texttt{ExactFAS} returns a FAS of a directed graph $\mathcal{G}$, which is a set of edges that can be reversed to make $\mathcal{G}$ acyclic and of maximal weight.
    If applied to a maximally consistent semi-complete partially directed graph $\mathcal{S}$, it returns a set $\mathbb{A}_i$ for $\mathcal{S}$ with respect to the consistency matrix $W$. If $\mathbb{A}_i$ is optimal, meaning that leads to a maximally consistent acyclic tournament, then it holds that:
    $$ \mathcal{T}_i = (\mathbb{V}, (\mathbb{E}_{\mathcal{S}} \setminus \mathbb{A}_i) \cup \mathbb{A}_i^T)$$
    where $\mathbb{A}_i^T$ is the transpose of $\mathbb{A}_i$.
    $$ \texttt{score}(\mathcal{T}_i) = \texttt{maxScore}$$
\end{itemize}

\paragraph{Soundness.} \emph{Every tournament $\mathcal{T}$ in \texttt{Results} is a maximally consistent acyclic tournament.}
\begin{itemize}
    \item {\bf Initialization:} The algorithm starts by computing the maximally consistent semi-complete partially directed graph $\mathcal{S}$.
    Then, it computes a FAS $\mathbb{A}_0$ of $\mathcal{S}$ with respect to the consistency matrix $W$. 
    By definition of \texttt{ExactFAS}, a FAS $\mathbb{A}_0$ is a set of edges that can be reversed to transform $\mathcal{S}$ into a maximally weighted acyclic tournament, $\mathcal{T}^0$;
    the tournament is then added to \texttt{Results} and the maximal consistency score $\texttt{maxScore}$ is computed.
    \item {\bf Iteration:} At every iteration, we compute a FAS of the semi-complete partially directed graph $\mathcal{S}$ with respect to the consistency matrix $W'$ in which some edges have been excluded from the solution space.
    a FAS $A$ is computed by \texttt{ExactFAS}$(\mathbb{E}_{\mathcal{S}}, W')$.
    The acyclic tournament $\mathcal{T}_{A}$ and its score are then computed.
    If the score of $\mathcal{T}_{A}$ is equal to $\texttt{maxScore}$, it is maximally consistent, thus it is added to \texttt{Results}.
\end{itemize}

\paragraph{Completeness.} \emph{Every maximally consistent acyclic tournament $\mathcal{T}$ is in \texttt{Results}.}

The solution space is explored by exclusion. The idea is that any maximally consistent solution is unique, thus removing from the solution space subsets of optimal FAS will force \texttt{ExactFAS} to search for other maximally consistent solutions.
This is achieved by iterating over the power set of the optimal FAS, contained in \texttt{maximalFAS}.
To exclude a set of edges $\mathbb{F}$ from the admissible solutions, we set the cost of reversing edges in $\mathbb{F}$ to $-\infty$, making them suboptimal by construction.
So we can build $W'$ as a copy of $W$ where $W[j, i] = -\infty$ for every edge $(i, j) \in \mathbb{F}$.
If there is another maximally consistent solution, $A'$, such that $\mathbb{F} \not \subset \mathbb{A}'$, then it will be found in the next iteration of the algorithm by \texttt{ExactFAS}$(\mathcal{S}, W')$.
We define $\mathbb{F}$ as the union of subsets of optimal solutions, $\mathbb{A}_i \in \texttt{maximalFAS}$. 
$$\mathbb{F} = \mathbb{F}_0 \cup \mathbb{F}_1 \cup \ldots \cup \mathbb{F}_k$$
where $\mathbb{F}_i \subset \mathbb{A}_i$. 
Given an undiscovered set $\mathbb{F}$ of forbidden edges:
\begin{itemize}
    \item $\mathbb{F} = \emptyset$, which corresponds to the case where the algorithm has not yet explored any subset of the optimal FAS.
    The algorithm will compute the first FAS $\mathbb{A}_0$ and add it to \texttt{Results}.
    \item $\mathbb{F} \neq \emptyset$, meaning that the algorithm has already explored some subsets of the optimal FAS.
    In this case, the algorithm will compute a FAS $\mathbb{A}_i$ that is either: 1) optimal, meaning that it is a maximally consistent acyclic tournament, or 2) suboptimal, meaning that it is not a maximally consistent acyclic tournament.
    In the first case, $\mathbb{A}_i$ will be added to $\texttt{Results}$; moreover, we will add to the \texttt{Queue} the combinatorial union between $\mathbb{F}$ and all the subsets of $\mathbb{A}_i$, which represents a tighter constraint on the admissible solutions.
    Eventually, this will lead to the discovery of a new maximally consistent acyclic tournament.
\end{itemize}

\paragraph{Termination.}  \emph{The MATS algorithm terminates.}

MATS systematically explores the power set of the optimal FAS, which is finite. 
Thus, the algorithm will eventually terminate.
Additionally, the algorithm adopts a caching mechanism to avoid exploring the same subsets multiple times.
Also, any set leading to a suboptimal solution allows us to identify sets of edges that can be excluded from the search space.
Indeed, any set containing a subset of edges that leads to a suboptimal solution will not be explored.

\end{proof}

\begin{proof}[Proposition~\ref{prop:5}]
    Following Proposition~\ref{prop:1}, the maximally consistent semi-complete PDG, $\mathcal{S}$, does not contain directed cycles.
    Moreover, since the consistency matrix is provided by a consistent knowledge base, it holds that for every pair of variables $X_i$ and $X_j$, if $X_i \succ X_j$ then $C_{X_i \succ X_j} > C_{X_j \succ X_i}$.
    The MATS algorithm will find all acyclic orientations of the undirected edges in $\mathcal{S}$ that do not introduce new cycles. 
    Among these, at least one will be the true causal order.
    Note that in the limit case where every edge is undirected, MATS will return the list of all acyclic tournaments.
\end{proof}

\begin{proof}[Proposition~\ref{prop:6}]
    The proof follows from Proposition~\ref{prop:4}. A strictly consistent expert guarantees that for every pair of variables $X_i$ and $X_j$, such that $X_i \succ X_j$, it holds that $C_{X_i \succ X_j} > C_{X_j \succ X_i}$.
    Meaning that the maximally consistent semi-complete partially directed graph $\mathcal{S}$ is an acyclic tournament, since there cannot be undirected edges.
\end{proof}

\begin{figure}
    \centering
    \begin{tikzpicture}
        \node[minimum height=0.5cm,minimum width=0.5cm] (Y) at (0, -2) {$Y$};
        \node[minimum height=0.5cm,minimum width=0.5cm] (X) at (-2, -2) {$X$};
        \node[minimum height=0.5cm,minimum width=0.5cm] (Z) at (-1, 0) {$Z$};

        \draw[->] (X) -- (Y);
        \draw[-] (Z) -- (Y);
        \draw[-,>=latex,ultra thick] (X) -- (Z);

        \node[minimum height=0.5cm,minimum width=0.5cm] (Y1) at (3, -2) {$Y$};
        \node[minimum height=0.5cm,minimum width=0.5cm] (X1) at (1, -2) {$X$};
        \node[minimum height=0.5cm,minimum width=0.5cm] (Z1) at (2, 0) {$Z$};

        \draw[->] (X1) -- (Y1);
        \draw[->] (Z1) -- (Y1);
        \draw[-,>=latex,ultra thick] (X1) -- (Z1);
    \end{tikzpicture}
    \caption{Non-identifiable subgraphs. In bold, the undirected edges are related to the treatment variable.} 
    \label{fig:non_identifiable_subgraphs}

\end{figure}


\section{Implementation Details}
\label{appendix:implementation}
\begin{algorithm}
    \caption{Maximally Weighted Acyclic Tournaments Search (MATS)}\label{algo:1}
    \textbf{Input}: $\mathbb{V}$, variables; $W$, weights\\
    \textbf{Output}: $Results$, a set of maximally weighted acyclic tournaments
    \begin{algorithmic}[1]
        \STATE $\mathcal{S} \gets \textsc{MaximallyWeightedGraph}(\mathbb{V}, W)$ 
        \STATE $(\mathbb{V}, \mathbb{E}_{\mathcal{S}}) \gets \mathcal{S}$
        \STATE $\mathbb{A} \gets \textsc{ExactFAS}(\mathbb{E}_{\mathcal{S}}, W)$ \COMMENT{Find feedback arc set}
        \STATE $\mathbb{E}_{\mathcal{T}} \gets (\mathbb{E}_{\mathcal{S}} \setminus \mathbb{A}) \cup \mathbb{A}^T$        
        \STATE $maxScore \gets \sum_{(i, j) \in \mathbb{E}_{\mathcal{T}}} W[i, j]$
        \STATE $Queue \gets \mathcal{P}(\mathbb{A})$ \COMMENT{Power set of $\mathbb{A}$}
        \STATE $maximalFAS \gets \{\mathbb{A}\}$
        \STATE $Results \gets \{\mathbb{E}_{\mathcal{T}}\}$
        \STATE $Cache \gets \emptyset$
        \WHILE{$Queue \neq \emptyset$}
            \STATE $\mathbb{F} \gets \textsc{Pop}(Queue)$
            \STATE $Cache \gets Cache \cup \mathbb{F}$
            \STATE $W' \gets \textsc{Copy}(W)$
            \FOR{$(i, j) \in \mathbb{F}$}
                \STATE $W'[j, i] \gets -\infty$
            \ENDFOR
            \STATE $\mathbb{A} \gets \textsc{ExactFAS}(\mathbb{E}_{\mathcal{S}}, W')$
            \STATE $\mathbb{E}_{\mathcal{T}} \gets (\mathbb{E}_{\mathcal{S}} \setminus \mathbb{A}) \cup \mathbb{A}^T$    
            \STATE $newScore \gets \sum_{(i, j) \in \mathbb{E}_{\mathcal{T}}} W[i, j]$
            \IF{$newScore = maxScore$ \textbf{and} $\mathbb{A} \notin maximalFAS$}
                \STATE $maximalFAS \gets maximalFAS \cup \{\mathbb{A}\}$
                \STATE $Results \gets Results \cup \{\mathbb{E}_{\mathcal{T}}\}$
                \FOR{$\mathbb{F}' \in \mathcal{P}(\mathbb{A})$}
                    \IF{$\mathbb{F}' \cup \mathbb{F} \notin Cache$ \textbf{and} $\mathbb{F}' \cup \mathbb{F} \notin Queue$}
                        \STATE \textsc{Push}(Queue, $\mathbb{F}' \cup \mathbb{F}$)
                    \ENDIF
                \ENDFOR
            \ELSE
                \FOR{$\mathbb{F}' \in Queue$}
                    \IF{$\mathbb{F} \subset \mathbb{F}'$}
                        \STATE \textsc{Remove}(Queue, $\mathbb{F}'$)
                    \ENDIF
                    \STATE $Cache \gets Cache \cup \{\mathbb{F}'\}$
                \ENDFOR
            \ENDIF
        \ENDWHILE
        \STATE \textbf{return} $Results$
    \end{algorithmic}
    \label{alg:mats}
\end{algorithm}

\paragraph{\bf Undirected Edges.} To reduce computational time, undirected edges are handled separately.
Before applying Algorithm~\ref{alg:mats}, all undirected edges are removed from the graph $\mathcal{S}$.
After generating a class of acyclic tournaments, each undirected edge is reintroduced as a directed one, oriented in one of its possible directions, as long as it doesn't create new cycles.
Since undirected edges contribute equally to consistency, their orientation does not change the consistency score of the maximal acyclic tournament, but reduces the size of the SCCs processed by \texttt{ExactFAS}.

\paragraph{\bf Data Generation.} For each node $x$ in the DAG,
\[x = f_x(Parents(x)) + \epsilon_x,\]
where $\epsilon_x$ is sampled from a Gaussian distribution, $Parents(x)$ is the set of parents of $x$ in the graph;
$f_x$ in the linear case is of the form $f_x = \sum_{i=1}^{k} x_i m_i$, whereas in the non-linear one is a function randomly picked from $\{sin(.), cos(.), square(.)\}$. 

\section{Additional Results}
\label{appendix:additional} 

\begin{figure}
    \includegraphics[width=\linewidth]{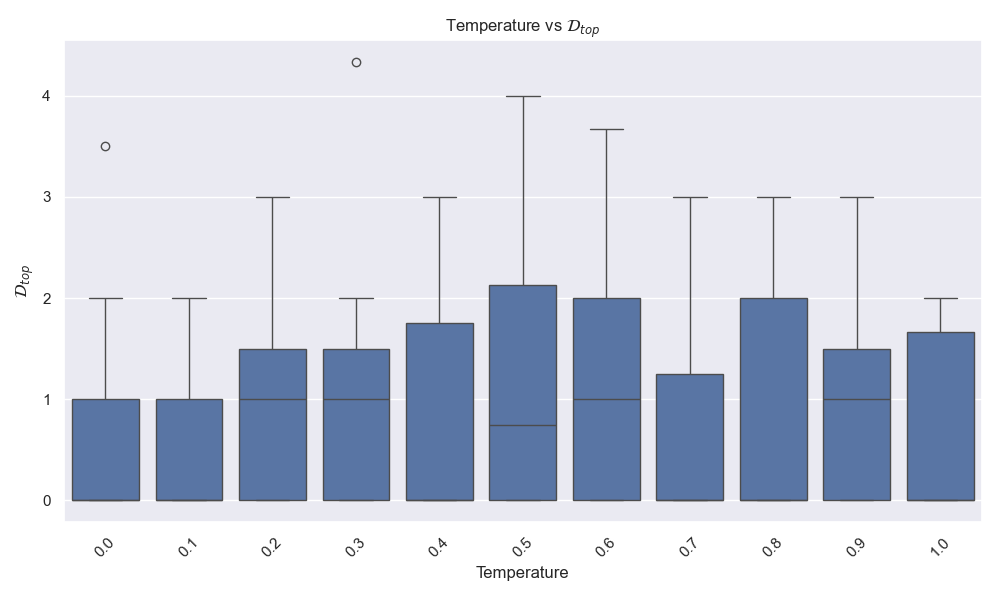}
    \caption{Boxplot intraclass $\mathcal{D}_{top}$ of MATS using \texttt{gpt-4.1-nano} with different temperatures. Error is bounded to values higher than 0. Note that when the median is not visible signifies it is equal to 0.}
    \label{fig:temperatures}
\end{figure}

This section provides further experimental results and analysis that complement those presented in the main text.
\begin{itemize}
    \item Figure~\ref{fig:temperatures} shows the results of MATS using \texttt{gpt-4.1-nano} with different temperatures.
    We can see that a lower temperature leads to a lower $\mathcal{D}_{top}$. 
    Indeed, a lower temperature leads to more deterministic outputs, reducing hallucinations.
    \item Tables~\ref{table:shd_linear} complements results in Table~\ref{tab:results} by providing SHD results on linear data.
    As argued in the main text, MATS provides a fully connected DAG, which leads to higher SHD values compared to data-driven baselines that return sparser graphs.
    \item Tables~\ref{table:shd_nonlinear} provides the SHD associated with non-linear data for data-driven methods. We can see how MATS does not consistently outperform other methods, mainly due to the fact that we are recovering tournaments.
    \item Tables~\ref{tab:d_top_linear_hybrid} and~\ref{tab:shd_linear_hybrid} present results of the hybridized approach using PC to estimate the skeleton, while using the approximated orders from MATS and Triplets to orient the edges.
    In this context, we can see a general improvement of the error associated with the estimated DAG.
    \item Table~\ref{tab:d_top_notears} and~\ref{tab:shd_notears} compare MATS to NOTEARS. In this context, we display both the error associated with the DAG discovered by NOTEARS and the relative CPDAG.
    \item Table~\ref{tab:results_llms_orders} showcases the best results obtained for intraclass $\mathcal{D}_{top}$ using different LLMs to compute the consistency matrix. 
    We can see that generally, \texttt{gpt-4.1-nano} and \texttt{mistral} perform similarly, followed by \texttt{llama3.1}.
    Most importantly, in many instances, MATS recovers only correct orders.
\end{itemize}

\begin{table}
\centering 
\scriptsize
\setlength{\tabcolsep}{2pt}
\begin{tabular}{cccccccccc}
\hline
\multicolumn{1}{|c|}{\multirow{2}{*}{\textbf{Graph}}} & \multicolumn{1}{c|}{\multirow{2}{*}{\textbf{PC}}}                                   & \multicolumn{1}{c|}{\multirow{2}{*}{\textbf{GES}}}                                  & \multicolumn{1}{c|}{\multirow{2}{*}{\textbf{NOTEARS}}}                            & \multicolumn{1}{c|}{\multirow{2}{*}{\textbf{BOSS}}}                                  & \multicolumn{1}{c|}{\multirow{2}{*}{\textbf{GRaSP}}}                                & \multicolumn{1}{c|}{\multirow{2}{*}{\textbf{GOBNILP}}} & \multicolumn{1}{c|}{\multirow{2}{*}{\textbf{Triplet}}} & \multicolumn{2}{c|}{\textbf{MATS (Our)}}                                              \\ \cline{9-10} 
\multicolumn{1}{|c|}{}                                & \multicolumn{1}{c|}{}                                                               & \multicolumn{1}{c|}{}                                                               & \multicolumn{1}{c|}{}                                                             & \multicolumn{1}{c|}{}                                                                & \multicolumn{1}{c|}{}                                                               & \multicolumn{1}{c|}{}                                  & \multicolumn{1}{c|}{}                                  & \multicolumn{1}{c|}{\textbf{MPDAG}}     & \multicolumn{1}{c|}{\textbf{$\pi$}} \\
\end{tabular}
\caption{SHD ($\downarrow$) of the estimated causal orders with linear synthetic data. Best results are highlighted in gray.}
\label{table:shd_linear}
\end{table}

\begin{table}
\centering 
\scriptsize
\setlength{\tabcolsep}{2pt}
\begin{tabular}{cccccccccc}
\hline
\multicolumn{1}{|c|}{\multirow{2}{*}{\textbf{Graph}}} & \multicolumn{1}{c|}{\multirow{2}{*}{\textbf{PC}}}                                    & \multicolumn{1}{c|}{\multirow{2}{*}{\textbf{GES}}}                                   & \multicolumn{1}{c|}{\multirow{2}{*}{\textbf{NOTEARS}}} & \multicolumn{1}{c|}{\multirow{2}{*}{\textbf{BOSS}}} & \multicolumn{1}{c|}{\multirow{2}{*}{\textbf{GRaSP}}} & \multicolumn{1}{c|}{\multirow{2}{*}{\textbf{GOBNILP}}} & \multicolumn{1}{c|}{\multirow{2}{*}{\textbf{Triplet}}}                              & \multicolumn{2}{c|}{\textbf{MATS (Our)}}                                                                                                                                  \\ \cline{9-10} 
\multicolumn{1}{|c|}{}                                & \multicolumn{1}{c|}{}                                                                & \multicolumn{1}{c|}{}                                                                & \multicolumn{1}{c|}{}                                  & \multicolumn{1}{c|}{}                               & \multicolumn{1}{c|}{}                                & \multicolumn{1}{c|}{}                                  & \multicolumn{1}{c|}{}                                                               & \multicolumn{1}{c|}{\textbf{MPDAG}}                                                 & \multicolumn{1}{c|}{\textbf{Causal Orders}}                                         \\
\end{tabular}
\caption{SHD ($\downarrow$) of the estimated causal orders with non-linear synthetic data. Best results are highlighted in gray.}
\label{table:shd_nonlinear}
\end{table}

\begin{table}
\centering
\scriptsize
\setlength{\tabcolsep}{2pt}
\begin{tabular}{ccccccccc}
\hline
\multicolumn{1}{|c|}{\multirow{2}{*}{\textbf{Graph}}} & \multicolumn{1}{c|}{\multirow{2}{*}{\textbf{PC}}}                                   & \multicolumn{1}{c|}{\multirow{2}{*}{\textbf{GES}}}                                  & \multicolumn{1}{c|}{\multirow{2}{*}{\textbf{NOTEARS}}} & \multicolumn{1}{l|}{\multirow{2}{*}{\textbf{BOSS}}}                                 & \multicolumn{1}{l|}{\multirow{2}{*}{\textbf{GRaSP}}}                                & \multicolumn{1}{l|}{\multirow{2}{*}{\textbf{GOBNILP}}} & \multicolumn{1}{c|}{\multirow{2}{*}{\textbf{Triplet + PC}}}                         & \multicolumn{1}{c|}{\multirow{2}{*}{\textbf{MATS + PC (Ours)}}}                     \\
\multicolumn{1}{|c|}{}                                & \multicolumn{1}{c|}{}                                                               & \multicolumn{1}{c|}{}                                                               & \multicolumn{1}{c|}{}                                  & \multicolumn{1}{l|}{}                                                               & \multicolumn{1}{l|}{}                                                               & \multicolumn{1}{l|}{}                                  & \multicolumn{1}{c|}{}                                                               & \multicolumn{1}{c|}{}                                                               \\
\end{tabular}
\caption{$\mathcal{D}_{top}$ ($\downarrow$) of the estimated causal graphs (obtained hybridizing MATS and Triplets with PC) with linear synthetic data. Best results are highlighted in gray.}
\label{tab:d_top_linear_hybrid}
\end{table}

\begin{table}
\centering
\scriptsize
\setlength{\tabcolsep}{2pt}
\begin{tabular}{ccccccccc}
\hline
\multicolumn{1}{|c|}{\textbf{Graph}}       & \multicolumn{1}{c|}{\textbf{PC}}                                                    & \multicolumn{1}{c|}{\textbf{GES}}                                                   & \multicolumn{1}{c|}{\textbf{NOTEARS}}                                               & \multicolumn{1}{c|}{\textbf{BOSS}}                                                  & \multicolumn{1}{c|}{\textbf{GRaSP}}                                                 & \multicolumn{1}{c|}{\textbf{GOBNILP}}   & \multicolumn{1}{c|}{\textbf{Triplet + PC}}                                          & \multicolumn{1}{c|}{\textbf{MATS + PC (Our)}}                                       \\ \hline
\multicolumn{1}{|c|}{\textbf{Asia}}        & \multicolumn{1}{c|}{$4.39^{\pm 2.34}$}                                              & \multicolumn{1}{c|}{$3.00^{\pm 1.73}$}                                              & \multicolumn{1}{c|}{$5.83^{\pm1.84}$}                                               & \multicolumn{1}{c|}{$3.00^{\pm 1.73}$}                                              & \multicolumn{1}{c|}{$3.00^{\pm 1.73}$}                                              & \multicolumn{1}{c|}{$14.50^{\pm 2.41}$} & \multicolumn{1}{c|}{$2.83^{\pm 0.69}$}                                              & \multicolumn{1}{c|}{\cellcolor[HTML]{C0C0C0}$2.22^{\pm 1.81}$} \\ \hline
\multicolumn{1}{|c|}{\textbf{Cancer}}      & \multicolumn{1}{c|}{\cellcolor[HTML]{C0C0C0}$0.00^{\pm 0.00}$} & \multicolumn{1}{c|}{\cellcolor[HTML]{C0C0C0}$0.00^{\pm 0.00}$} & \multicolumn{1}{c|}{$1.00^{\pm0.24}$}                                               & \multicolumn{1}{c|}{\cellcolor[HTML]{C0C0C0}$0.00^{\pm 0.00}$} & \multicolumn{1}{c|}{\cellcolor[HTML]{C0C0C0}$0.00^{\pm 0.00}$} & \multicolumn{1}{c|}{$7.50^{\pm 1.72}$}  & \multicolumn{1}{c|}{$4.20^{\pm 0.40}$}                                              & \multicolumn{1}{c|}{$0.33^{\pm 0.47}$}                                              \\
\end{tabular}
\caption{SHD ($\downarrow$) of the estimated causal graphs (obtained hybridizing MATS and Triplets with PC) with linear synthetic data. Best results are highlighted in gray.}
\label{tab:shd_linear_hybrid}
\end{table}

\begin{table}[]
\centering
\begin{tabular}{ccccccc}
\hline
\multicolumn{1}{|c|}{\multirow{3}{*}{\textbf{Graph}}} & \multicolumn{4}{c|}{\textbf{NOTEARS}}                                                                                                                                                                         & \multicolumn{2}{c|}{\textbf{MATS (Our)}}                                                                                                                                  \\ \cline{2-7} 
\multicolumn{1}{|c|}{}                                & \multicolumn{2}{c|}{Linear}                                                                                                  & \multicolumn{2}{c|}{Non-Linear}                                                & \multicolumn{1}{c|}{\multirow{2}{*}{\textbf{MPDAG}}}                                & \multicolumn{1}{c|}{\multirow{2}{*}{\textbf{$\pi$}}}                                \\ \cline{2-5}
\multicolumn{1}{|c|}{}                                & \multicolumn{1}{c|}{DAG}                                                            & \multicolumn{1}{c|}{CPDAG}             & \multicolumn{1}{c|}{DAG}              & \multicolumn{1}{c|}{CPDAG}             & \multicolumn{1}{c|}{}                                                               & \multicolumn{1}{c|}{}                                                               \\ \hline
\multicolumn{1}{|c|}{\textbf{Asia}}                   & \multicolumn{1}{c|}{\cellcolor[HTML]{C0C0C0}$0.75^{\pm 0.97}$} & \multicolumn{1}{c|}{$4.08^{\pm 0.92}$} & \multicolumn{1}{c|}{$7.40^{\pm1.50}$} & \multicolumn{1}{c|}{$6.06^{\pm0.36}$}  & \multicolumn{1}{c|}{$2.87^{\pm 1.81}$}                                              & \multicolumn{1}{c|}{$1.22^{\pm 1.03}$}                                              \\
\end{tabular}
\caption{$\mathcal{D}_{top}$ ($\downarrow$) comparing MATS to NOTEARS using linear and non-linear data. We display both the $\mathcal{D}_{top}$ associated with the DAG  found by NOTEARS and the relative CPDAG. Best results are highlighted in gray.}
\label{tab:d_top_notears}
\end{table}

\begin{table}[]
\centering
\begin{tabular}{ccccccc}
\hline
\multicolumn{1}{|c|}{\multirow{3}{*}{\textbf{Graph}}} & \multicolumn{4}{c|}{\textbf{NOTEARS}}                                                                                                                                                                                                                         & \multicolumn{2}{c|}{\textbf{MATS (Our)}}                                                                    \\ \cline{2-7} 
\multicolumn{1}{|c|}{}                                & \multicolumn{2}{c|}{Linear}                                                                                                                                              & \multicolumn{2}{c|}{Non-Linear}                                                    & \multicolumn{1}{c|}{\multirow{2}{*}{\textbf{MPDAG}}} & \multicolumn{1}{c|}{\multirow{2}{*}{\textbf{$\pi$}}} \\ \cline{2-5}
\multicolumn{1}{|c|}{}                                & \multicolumn{1}{c|}{DAG}                                                            & \multicolumn{1}{c|}{CPDAG}                                                         & \multicolumn{1}{c|}{DAG}                 & \multicolumn{1}{c|}{CPDAG}              & \multicolumn{1}{c|}{}                                & \multicolumn{1}{c|}{}                                \\ \hline
\multicolumn{1}{|c|}{\textbf{Asia}}                   & \multicolumn{1}{c|}{\cellcolor[HTML]{C0C0C0}$2.50^{\pm 1.58}$} & \multicolumn{1}{c|}{$5.83^{\pm1.84}$}                                              & \multicolumn{1}{c|}{$13.40^{\pm 1.96}$}  & \multicolumn{1}{c|}{$15.83^{\pm 0.33}$} & \multicolumn{1}{c|}{$24.00^{\pm2.25}$}               & \multicolumn{1}{c|}{$22.00^{\pm1.63}$}               \\
\end{tabular}
\caption{SHD ($\downarrow$) comparing MATS to NOTEARS using linear and non-linear data. We display both the $\mathcal{D}_{top}$ associated with the DAG found by NOTEARS and the relative CPDAG. Best results are highlighted in gray.}
\label{tab:shd_notears}
\end{table}

\begin{table}
    \footnotesize
    \centering 
\begin{tabular}{cccc}
\hline
\multicolumn{1}{|c|}{\textbf{Graph}}  & \multicolumn{1}{c|}{\texttt{gpt-4.1-nano}}      & \multicolumn{1}{c|}{\texttt{mistral:7b}}        & \multicolumn{1}{c|}{\texttt{llama3.1}}          \\ \hline
\multicolumn{1}{|c|}{\textbf{Asia}}        & \multicolumn{1}{c|}{\cellcolor[HTML]{C0C0C0}$0.00 \pm 0.00$} & \multicolumn{1}{c|}{\cellcolor[HTML]{C0C0C0}$0.00 \pm 0.00$} & \multicolumn{1}{c|}{$2.00 \pm 0.00$}                         \\ \hline
\multicolumn{1}{|c|}{\textbf{Cancer}}      & \multicolumn{1}{c|}{\cellcolor[HTML]{C0C0C0}$0.00 \pm 0.00$} & \multicolumn{1}{c|}{\cellcolor[HTML]{C0C0C0}$0.00 \pm 0.00$} & \multicolumn{1}{c|}{$2.00 \pm 0.00$}                         \\ \hline
\multicolumn{1}{|c|}{\textbf{Climate}}     & \multicolumn{1}{c|}{\cellcolor[HTML]{C0C0C0}$0.00 \pm 0.00$} & \multicolumn{1}{c|}{\cellcolor[HTML]{C0C0C0}$0.00 \pm 0.00$} & \multicolumn{1}{c|}{$3.00 \pm 0.00$}                         \\ \hline
\multicolumn{1}{|c|}{\textbf{Covid 1}}     & \multicolumn{1}{c|}{\cellcolor[HTML]{C0C0C0}$0.00 \pm 0.00$} & \multicolumn{1}{c|}{\cellcolor[HTML]{C0C0C0}$0.00 \pm 0.00$} & \multicolumn{1}{c|}{\cellcolor[HTML]{C0C0C0}$0.00 \pm 0.00$} \\ \hline
\multicolumn{1}{|c|}{\textbf{Covid 2}}     & \multicolumn{1}{c|}{\cellcolor[HTML]{C0C0C0}$0.00 \pm 0.00$} & \multicolumn{1}{c|}{\cellcolor[HTML]{C0C0C0}$0.00 \pm 0.00$} & \multicolumn{1}{c|}{\cellcolor[HTML]{C0C0C0}$0.00 \pm 0.00$} \\ \hline
\multicolumn{1}{|c|}{\textbf{Covid 3}}     & \multicolumn{1}{c|}{$0.50 \pm 0.50$}                         & \multicolumn{1}{c|}{$0.50 \pm 0.50$}                         & \multicolumn{1}{c|}{\cellcolor[HTML]{C0C0C0}$0.00 \pm 0.00$} \\ \hline
\multicolumn{1}{|c|}{\textbf{Covid 4}}     & \multicolumn{1}{c|}{\cellcolor[HTML]{C0C0C0}$0.00 \pm 0.00$} & \multicolumn{1}{c|}{\cellcolor[HTML]{C0C0C0}$0.00 \pm 0.00$} & \multicolumn{1}{c|}{$1.00 \pm 0.00$}                         \\
\end{tabular}
    \caption{Best intraclass $\mathcal{D}_{top}$ ($\downarrow$) of MATS for causal orders using different LLMs. In gray best performing method. '-' indicates that the method did not return a valid causal order.}
    \label{tab:results_llms_orders}
\end{table}

\pagebreak
\section{Prompts \& Graphs}
\label{appendix:graphs}

\begin{table}
    \centering
    \begin{tabular}{|l|c|c|}
    \hline
    \multicolumn{1}{|c|}{\textbf{Graph}} & \textbf{Nodes} & \textbf{Edges} \\ \hline
    Asia~\ref{fig:asia}              & 8              & 8              \\ \hline
    Cancer~\ref{fig:cancer}          & 5              & 4              \\ \hline
    Climate~\ref{fig:climate}             & 8              & 8              \\ \hline
    Covid 1~\ref{fig:covid_1}           & 3              & 2              \\ \hline
    Covid 2~\ref{fig:covid_2}            & 4              & 5              \\ \hline
    Covid 3~\ref{fig:covid_3}            & 4              & 3              \\ \hline
    Covid 4~\ref{fig:covid_4}           & 4              & 6              \\ \hline
    Genetic~\ref{fig:genetic}              & 6              & 5              \\ \hline
    MSU~\ref{fig:msu}              & 5              & 6              \\ \hline
    Neighborhood~\ref{fig:neighborhood}       & 5              & 8              \\ \hline
    Sachs~\ref{fig:sachs}                & 11             & 17             \\ \hline
    Supermarket~\ref{fig:supermarket}         & 7              & 12             \\ \hline
\end{tabular}
\caption{Causal DAGs used in the experiments.}

\end{table}

  
    \caption{Prompt used to query the LLM for rephrasing.}
\end{figure}

\begin{figure}
    \centering
    %
  
    \caption{Prompt used to query the LLM for causal relationships.}
\end{figure}

\begin{figure*}
    \centering
    \begin{tikzpicture}
        \node[draw] (A) at (0,0) {Visiting Asia};
        \node[draw] (B) at (0,-1) {Tuberculosis};
        \node[draw] (C) at (3,-1) {Lung Cancer};
        \node[draw] (H) at (1, -2) {Either Tuberculosis or Lung Cancer};
        \node[draw] (D) at (5,-2) {Bronchitis};
        \node[draw] (E) at (4,0) {Smoker};
        \node[draw] (F) at (1,-3) {X-ray};
        \node[draw] (G) at (5,-3) {Dyspnea};

        \draw[->,>=latex] (A) -- (B);
        \draw[->,>=latex] (B) -- (H);
        \draw[->,>=latex] (C) -- (H);
        \draw[->,>=latex] (E) -- (C);
        \draw[->,>=latex] (H) -- (F);
        \draw[->,>=latex] (C) -- (D);
        \draw[->,>=latex] (D) -- (G);
        \draw[->,>=latex] (H) -- (G);

    \end{tikzpicture}
    \caption{Asia causal DAG from \texttt{bnlearn}.}
\label{fig:asia}
\end{figure*}   
\begin{figure*}
    \centering 
    \begin{tikzpicture}
        \node[draw] (A) at (0,0) {Pollution};
        \node[draw] (B) at (1.5,-1) {Cancer};
        \node[draw] (C) at (3,0) {Smoking};
        \node[draw] (D) at (0, -2) {X-ray};
        \node[draw] (E) at (3,-2) {Dyspnea};

        \draw[->,>=latex] (A) -- (B);
        \draw[->,>=latex] (C) -- (B);
        \draw[->,>=latex] (B) -- (D);
        \draw[->,>=latex] (B) -- (E);
    \end{tikzpicture}
    \caption{Cancer causal DAG from \texttt{bnlearn}.}
\label{fig:cancer}
\end{figure*}
\begin{figure*}
    \centering
    \begin{tikzpicture}
        \node[draw] (A) at (0,0) {Temperature};
        \node[draw] (B) at (0,-2) {Dew Point};
        \node[draw] (C) at (2,-1) {Relative Humidity};
        \node[draw] (D) at (6,-1) {Transmission Rate};
        \node[draw] (E) at (10,-1) {Incidence Rate};
        \node[draw] (F) at (10,0) {Susceptible Population};
        \node[draw] (G) at (10,-2) {Infected Population};
        \node[draw] (H) at (14,-1) {Observed Incidence Rate};

        \draw[->,>=latex] (A) -- (C);
        \draw[->,>=latex] (B) -- (C);
        \draw[->,>=latex] (C) -- (D);
        \draw[->,>=latex] (A) -- (D);
        \draw[->,>=latex] (D) -- (E);
        \draw[->,>=latex] (F) -- (E);
        \draw[->,>=latex] (G) -- (E);
        \draw[->,>=latex] (E) -- (H);
    \end{tikzpicture}
    \caption{Climate causal DAG.}
    \label{fig:climate}
\end{figure*}
\begin{figure*}
    \centering
    \begin{tikzpicture}[{black, rectangle, draw, inner sep=0.1cm,minimum height=0.5cm,minimum width=0.5cm}]
        \node[draw] (A) at (0,0) {Age};
        \node[draw] (C) at (1.5,-1) {App usage};
        \node[draw] (B) at (3,0) {COVID-19 infection};
        
        \draw[->,>=latex] (A) -- (C);
        \draw[->,>=latex] (B) -- (C);	
        
    \end{tikzpicture}		
    \caption{Covid 1 causal DAG.}
    \label{fig:covid_1}
\end{figure*}
\begin{figure*}
    \centering
    \begin{tikzpicture}[{black, rectangle, draw, inner sep=0.1cm,minimum height=0.5cm,minimum width=0.5cm}]
    
    \node[draw] (A) at (0,0) {\footnotesize Smoking};
    \node[draw] (C) at (3,-1) {\footnotesize COVID-19 testing};
    \node[draw] (B) at (3,0) {\footnotesize COVID-19 severity};
    \node[draw] (D) at (0,-1) {\footnotesize Healthcare worker};
    
    \draw[->,>=latex] (D) -- (A);
    \draw[->,>=latex] (D) -- (C);	
    \draw[->,>=latex] (B) -- (C);	
    \end{tikzpicture}
    \caption{Covid 2 causal DAG.}
    \label{fig:covid_2}
\end{figure*}
\begin{figure*}
    \centering
    \begin{tikzpicture}[{black, rectangle, draw, inner sep=0.1cm,minimum height=0.5cm,minimum width=0.5cm}]
    
    \node[draw] (A) at (0,0) {ACE-inhibitors};
    \node[draw] (C) at (0,1) {Hospitalisation};
    \node[draw] (B) at (3,1) {Death};
    \node[draw] (D) at (0,2) {Frailty};
    
    \draw[->,>=latex] (A) -- (C);
    \draw[->,>=latex] (C) -- (B);	
    \draw[->,>=latex] (A) -- (B);	
    \draw[->,>=latex] (D) -- (C);	
    \draw[->,>=latex] (D) -- (B);	
    \end{tikzpicture}
    \caption{Covid 3 causal DAG.}
    \label{fig:covid_3}
\end{figure*}
\begin{figure*}
    \centering
    \begin{tikzpicture}[{black, rectangle, draw, inner sep=0.1cm,minimum height=0.5cm,minimum width=0.5cm}]
    
    \node[draw] (D) at (0,2) {\footnotesize Prevalence of diabetes} ;
    \node[draw] (Y) at (4,0) {\footnotesize Infection hospitalization rate};
    \node[draw] (X) at (0,0) {\footnotesize COVID-19 incidence};
    \node[draw, text width=3cm] (B) at (4,2) {\footnotesize Number of intensive care beds per inhabitant};
    \node[draw] (A) at (0,1) {\footnotesize Proportion of population over 60};
    
    \draw[->,>=latex] (D) -- (Y);
    \draw[->,>=latex] (X) -- (Y);
    \draw[->,>=latex] (A) -- (Y);
    \draw[->,>=latex] (B) -- (Y);
    \draw[->,>=latex] (A) -- (D);
    \draw[->,>=latex] (A) -- (X);
    
    \end{tikzpicture}
    \caption{Covid 4 causal DAG.}
    \label{fig:covid_4}
\end{figure*}
\begin{figure*}
    \centering
    \begin{tikzpicture}[{black, rectangle, draw, inner sep=0.1cm,minimum height=0.5cm,minimum width=0.5cm}]
    
        \node[draw] (G1) at (0,1) {\small Gene FTO};
        \node[draw] (G2) at (0,0) {\small Gene MC4R};
        \node[draw] (G3) at (0,-1) {\small Gene TMEM18};
        \node[draw] (G4) at (0,-2) {\small Gene GNPDA2};
        \node[draw] (F) at (3,0) {\small Fat mass};
        \node[draw] (B) at (3,-1) {\small Bone mineral density};
        
        \draw[->,>=latex] (G1) -- (F);
        \draw[->,>=latex] (G2) -- (F);
        \draw[->,>=latex] (G3) -- (F);
        \draw[->,>=latex] (G4) -- (F);
        \draw[->,>=latex] (F) -- (B);	
    \end{tikzpicture}		
    \caption{Genetic causal DAG.}
    \label{fig:genetic}
    \end{figure*}
    
\begin{figure*}
    \centering
    \begin{tikzpicture}[{black, rectangle, draw, inner sep=0.1cm,minimum height=0.5cm,minimum width=0.5cm}]
    
    \node[draw] (Z) at (0,0) {\footnotesize Time to thrombolysis};
    \node[draw] (X) at (-3,0) {\footnotesize Mobile stroke unit};
    \node[draw] (Y) at (3,0) {\footnotesize Functional outcome} ;
    \node[draw] (L1) at (-3,1) {\footnotesize Stroke severity} ;
    \node[draw] (L2) at (1,1) {\footnotesize Systolic blood pressure} ;
    
    \draw[->,>=latex] (X) -- (Z);
    \draw[->,>=latex] (Z) -- (Y);
    \draw[->,>=latex] (L1)  -- (Z);
    \draw[->,>=latex] (L1)  -- (Y);
    \draw[->,>=latex] (L2)  -- (Z);
    \draw[->,>=latex] (L2)  -- (Y);
    
    \end{tikzpicture}
    \caption{MSU causal DAG.}
    \label{fig:msu}

\end{figure*}
\begin{figure*}
    \centering
    \begin{tikzpicture}[{black, rectangle, draw, inner sep=0.1cm,minimum height=0.5cm,minimum width=0.5cm}]
    
    \node[draw] (Lo) at (0,0) {Low neighborhood socioeconomic position} ;
    \node[draw] (La) at (0,-1) {Lack of services};
    \node[draw] (E) at (8,1) {Exposure to high criminality levels};
    \node[draw] (R) at (8,0) {Reduced walking in one's neighborhood};
    \node[draw] (LoG) at (0,1) {Low accessibility to green spaces};
    \node[draw] (P) at (8,-1) {Proximity to polluting industry};
    
    \draw[->,>=latex] (LoG) -- (Lo);
    \draw[->,>=latex] (LoG) -- (R);
    \draw[->,>=latex] (P) -- (Lo);
    \draw[->,>=latex] (P) -- (R);
    \draw[->,>=latex] (Lo) -- (La);
    \draw[->,>=latex] (Lo) -- (E);
    \draw[->,>=latex] (La) -- (R);
    \draw[->,>=latex] (E) -- (R);
    \end{tikzpicture}
    \caption{Neighborhood causal DAG.}
    \label{fig:neighborhood}
\end{figure*}
\begin{figure*}
    \centering
    \begin{tikzpicture}
        \node[draw] (A) at (0,0) {PKC};
        \node[draw] (B) at (0,-2) {PKA};
        \node[draw] (C) at (-2,-4) {Raf};
        \node[draw] (D) at (-1,-6) {Mek};
        \node[draw] (E) at (-0.5,-8) {ERK};
        \node[draw] (F) at (0,-10) {Akt};
        \node[draw] (G) at (2,-4) {Jnk};
        \node[draw] (H) at (4,-4) {p38};
        \node[draw] (I) at (6,-0) {Plcg};
        \node[draw] (J) at (6,-4) {PIP2};
        \node[draw] (K) at (8,-2) {PIP3};

        \draw[->,>=latex] (A) -- (B);
        \draw[->,>=latex] (A) -- (C);
        \draw[->,>=latex] (A) -- (D);
        \draw[->,>=latex] (A) -- (G);
        \draw[->,>=latex] (A) -- (H);
        \draw[->,>=latex] (B) -- (E);
        \draw[->,>=latex] (B) -- (F);
        \draw[->,>=latex] (B) -- (C);
        \draw[->,>=latex] (B) -- (D);
        \draw[->,>=latex] (B) -- (G);
        \draw[->,>=latex] (B) -- (H);
        \draw[->,>=latex] (C) -- (D);
        \draw[->,>=latex] (D) -- (E);
        \draw[->,>=latex] (E) -- (F);
        \draw[->,>=latex] (I) -- (J);
        \draw[->,>=latex] (I) -- (K);
        \draw[->,>=latex] (K) -- (J);

    \end{tikzpicture}
    \caption{Sachs causal DAG from \texttt{bnlearn}.}
    \label{fig:sachs}
\end{figure*}
\begin{figure*}
    \centering
    \begin{tikzpicture}[{black, rectangle, draw, inner sep=0.1cm,minimum height=0.5cm,minimum width=0.5cm}]
    
    \node[draw] (N) at (1,2) {Neighborhood education and income} ;
    \node[draw] (E) at (-2,0) {Education};
    \node[draw] (I) at (1,1) {Income};
    \node[draw] (S) at (5,1) {Supermarket characteristics};
    \node[draw] (F) at (5,0) {Food preferences};
    \node[draw] (P) at (10,0) {Purchased food};
    \node[draw] (W) at (10,1) {Weights status and adiposity};
    
    \draw[->,>=latex] (E) -- (N);
    \draw[->,>=latex] (E) -- (I);
    \draw[->,>=latex] (E) -- (F);
    \draw[->,>=latex] (I) -- (N);
    \draw[->,>=latex] (I) -- (S);
    \draw[->,>=latex] (I) -- (P);
    \draw[->,>=latex] (F) -- (S);
    \draw[->,>=latex] (F) -- (P);
    \draw[->,>=latex] (N) -- (S);
    \draw[->,>=latex] (N) -- (W);
    \draw[->,>=latex] (S) -- (P);
    \draw[->,>=latex] (P) -- (W);
    
    \end{tikzpicture}
    \caption{Supermarket causal DAG.}
    \label{fig:supermarket}
\end{figure*}

\end{document}